\documentclass{article}

\usepackage[T1]{fontenc}    
\usepackage{helvet}         
\usepackage[margin=1in]{geometry}
\usepackage[numbers,sort&compress]{natbib}
\usepackage[
  colorlinks=false,
  linkbordercolor={0 1 0},
  citebordercolor={0 1 0},
  urlbordercolor={0 1 0},
  pdfborder={0 0 1}
]{hyperref} 
\makeatletter
\def\hyper@natlinkbreak#1#2{#1} 
\makeatother
\usepackage{url}            
\usepackage{graphicx}       
\usepackage{booktabs}       
\usepackage{amsfonts}       
\usepackage{nicefrac}       
\usepackage{microtype}      
\usepackage[table]{xcolor}  
\usepackage{multirow}       
\usepackage{listings}       
\usepackage[most]{tcolorbox} 
\usepackage{wrapfig}        
\usepackage[labelfont=bf]{caption} 
\usepackage{setspace}       
\usepackage{hyphenat}       
\usepackage{fancyhdr}       
\usepackage{fontawesome5}   
\newsavebox{\teaserbox}

\definecolor{rowbg}{RGB}{245,245,245}
\definecolor{PromptBg}{RGB}{250,250,250}
\definecolor{PromptRule}{RGB}{120,120,120}
\definecolor{XiaoxiTitleBg}{RGB}{242,249,255}
\definecolor{XiaoxiTitleFg}{HTML}{262130}

\definecolor{cvprblue}{rgb}{0.21,0.49,0.74}
\definecolor{HeaderGray}{gray}{0.90}   
\definecolor{StripeGray}{gray}{0.96}   
\definecolor{SpiderBlue}{RGB}{220,236,250}
\usepackage{graphicx}   
\usepackage{array}      
\usepackage{booktabs}   
\usepackage{longtable} 

\newcommand{\institutionlogofiles}{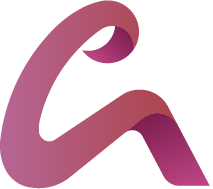,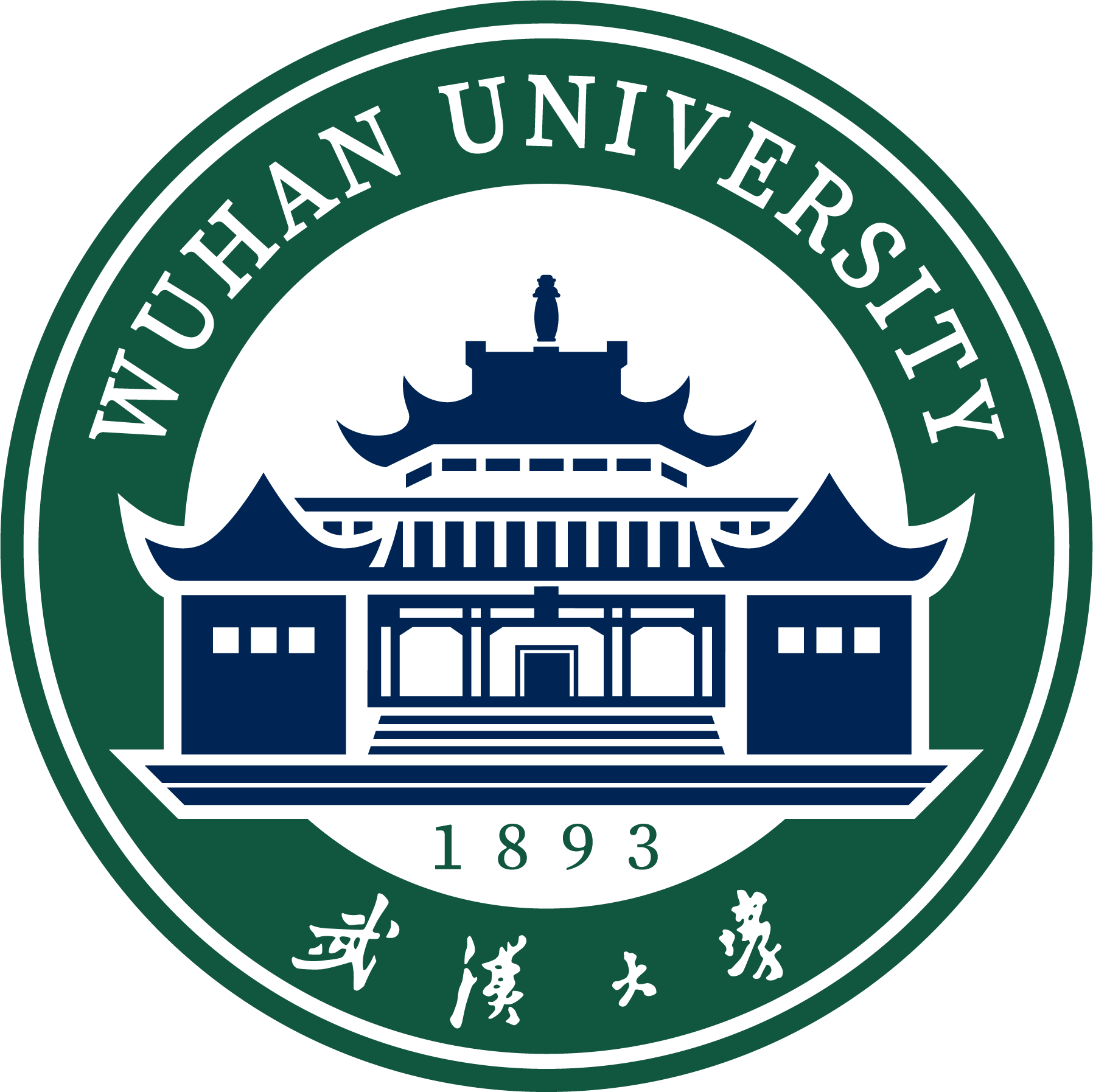,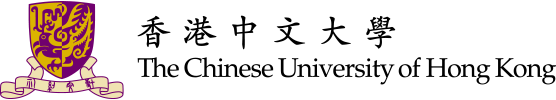}
\newcommand{\paperheadertitle}{From Web to Pixels: Bringing Agentic Search into Visual Perception}
\makeatletter
\newif\ifwe@institutionlogofound
\newcommand{\institutionlogo}{%
  \begingroup
  \we@institutionlogofoundfalse
  \@for\we@institutionlogofile:=\institutionlogofiles\do{%
    \IfFileExists{\we@institutionlogofile}{%
      \we@institutionlogofoundtrue
      \includegraphics[height=0.7cm]{\we@institutionlogofile}\hspace{0.5em}%
    }{}%
  }%
  \ifwe@institutionlogofound\else
    {\sffamily\bfseries Institution Logo}%
  \fi
  \endgroup
}
\makeatother

\fancypagestyle{plain}{%
  \fancyhf{}
  \fancyhead[L]{\institutionlogo}
  \fancyfoot[C]{\thepage}

}


\newtcblisting{promptbox}[1]{
  enhanced,
  breakable,
  listing only,
  listing engine=listings,
  title={#1},
  colback=PromptBg,
  colframe=PromptBg,
  coltitle=black,
  colbacktitle=PromptBg,
  fonttitle=\bfseries\footnotesize,
  boxrule=0pt,
  frame hidden,
  borderline west={0.8pt}{0pt}{PromptRule},
  sharp corners,
  left=1.4mm,
  right=0.8mm,
  top=0.4mm,
  bottom=0.5mm,
  toptitle=0.2mm,
  bottomtitle=0.2mm,
  lefttitle=0mm,
  before skip=0.45em,
  after skip=0.75em,
  listing options={
    basicstyle=\ttfamily\scriptsize,
    breaklines=true,
    breakatwhitespace=false,
    columns=fullflexible,
    keepspaces=true,
    showstringspaces=false
  }
}

\makeatletter
\def\we@abstract{}
\def\we@authornames{}
\def\we@affiliations{}
\def\we@contact{}
\def\we@corresponding{}
\def\we@projectwebsite{}
\def\we@github{}
\long\def\weabstract#1{\gdef\we@abstract{#1}}
\newcommand{\weauthors}[1]{\gdef\we@authornames{#1}}
\newcommand{\weaffiliations}[1]{\gdef\we@affiliations{#1}}
\newcommand{\wecontact}[1]{\gdef\we@contact{#1}}
\newcommand{\wecorresponding}[1]{\gdef\we@corresponding{#1}}
\newcommand{\weprojectwebsite}[1]{\gdef\we@projectwebsite{#1}}
\newcommand{\wegithub}[1]{\gdef\we@github{#1}}
\newcommand{\we@printauthors}{%
  {\normalsize\sffamily\bfseries \we@authornames\par}%
  \vskip 0.05cm
  {\fontsize{8.8}{11}\selectfont \we@affiliations\par}%
  \ifx\we@contact\@empty\else
    \vskip 0.05cm
    {\fontsize{8.8}{11}\selectfont\ttfamily \we@contact\par}%
  \fi
  \ifx\we@projectwebsite\@empty\else
    \vskip 0.05cm
    {\fontsize{8.8}{11}\selectfont \faGlobeAmericas\ Project Website: \url{\we@projectwebsite}\par}%
  \fi
  \ifx\we@github\@empty\else
    \vskip 0.05cm
    {\fontsize{8.8}{11}\selectfont \faGithub\ GitHub: \url{\we@github}\par}%
  \fi
}
\renewcommand{\maketitle}{%
  \par
  \begingroup
    \renewcommand{\thefootnote}{\fnsymbol{footnote}}%
    \renewcommand{\@makefnmark}{\hbox to \z@{$^{\@thefnmark}$\hss}}%
    \long\def\@makefntext##1{%
      \parindent 1em\noindent
      \hbox to 1.8em{\hss $\m@th ^{\@thefnmark}$}##1%
    }%
    \thispagestyle{plain}%
    \begin{tcolorbox}[
      enhanced,
      frame hidden,
      colback=XiaoxiTitleBg,
      left=0.5cm,
      right=0.5cm,
      top=0.5cm,
      bottom=0.5cm,
      arc=10pt,
      before skip=0pt,
      after skip=0.38cm,
      grow to left by=1.5pt,
      grow to right by=1.5pt
    ]
      \setlength{\parindent}{0pt}%
      \setlength{\parskip}{0.5cm}%
      {\setlength{\parskip}{0pt}%
       \raggedright
       \nohyphens
       {\setstretch{1.05}%
        {\fontsize{20.2}{21.5}\selectfont\fontfamily{phv}\bfseries\@title\par}}%
       \vskip 0.25cm
       \we@printauthors
      }%
      \ifx\we@abstract\@empty\else
        {\color{XiaoxiTitleFg}\we@abstract\par}%
      \fi
    \end{tcolorbox}%
    \ifx\we@corresponding\@empty\else
      \footnotetext[2]{\we@corresponding}%
    \fi
    \@thanks
  \endgroup
  \let\maketitle\relax
  \let\thanks\relax
}
\makeatother

\title{\paperheadertitle}

\weauthors{Bokang Yang\textsuperscript{1}, Xinyi Sun\textsuperscript{2}, Kaituo Feng\textsuperscript{3}, Xingping Dong\textsuperscript{2}, Dongming Wu\textsuperscript{3,$\dagger$},  Xiangyu Yue\textsuperscript{3,$\dagger$}}
\weaffiliations{\textsuperscript{1}Shenzhen Loop Area Institute \quad \textsuperscript{2}Wuhan University \quad \textsuperscript{3}CUHK MMLab }
\wecorresponding{Corresponding author}
\weprojectwebsite{https://pixel-searcher.github.io/}
\wegithub{https://github.com/yangbokang/Pixel-Searcher}

\begin{document}

\weabstract{%
Visual perception connects high-level semantic understanding to pixel-level perception, but most existing settings assume that the decisive evidence for identifying a target is already in the image or frozen model knowledge.
We study a more practical yet harder open-world case where a visible object must first be resolved from external facts, recent events, long-tail entities, or multi-hop relations before it can be localized.
We formalize this challenge as \textbf{Perception Deep Research} and introduce \textbf{WebEyes}, an object-anchored benchmark with verifiable evidence, knowledge-intensive queries, precise box/mask annotations, and three task views: Search-based Grounding, Search-based Segmentation, and Search-based VQA.
WebEyes contains 120 images, 473 annotated object instances, 645 unique QA pairs, and 1,927 task samples.
We further propose \textbf{Pixel-Searcher}, an agentic search-to-pixel workflow that resolves hidden target identities and binds them to boxes, masks, or grounded answers.
Experiments show that Pixel-Searcher achieves the strongest open-source performance across all three task views, while failures mainly arise from evidence acquisition, identity resolution, and visual instance binding.
}

\maketitle

\section{Introduction}
\label{sec:intro}


Visual perception is a foundation of multimodal intelligence, not only for recognizing visual entities but also for grounding language-level intent into boxes, masks, and region-level answers. Grounding and segmentation thus serve as key interfaces between semantic understanding and pixel-level perception. With the development of multimodal large language models (MLLMs)~\citep{feng2025video,shen2025vlm,yu2026latent,yu2026perception,wu2026reinforcing,fan2025sophiavl}, recent progress has pushed visual perception from visible-category recognition toward grounding implicit targets inferred from internal model knowledge~\citep{kamath2021mdetr,liu2024grounding,lai2024lisa,rasheed2024glamm}. Yet open-world settings introduce a more practical but harder case: \textit{The object may be visible, while the evidence needed to identify it lies outside the image and beyond frozen model knowledge}~\citep{wang2017fvqa,marino2019ok,jiang2024mmsearch,wu2025mmsearch}. 
Inspired by the recent progress of Deep Research~\citep{comanici2025gemini,fan2026exploring,singh2025openai} in knowledge-intensive tasks, we revisit visual perception from a broader perspective. 
Recognizing that real-world perception queries often involve up-to-date or knowledge-intensive information rather than direct visual attributes, we ask a natural question: \textit{can we build a visual perception search agent that actively performs multi-hop web search and reasoning to gather external knowledge for grounded visual perception?}

We formulate this setting as \textbf{Perception Deep Research}, where a model must resolve a target identity from external evidence and bind it to a concrete visual instance. Given an image and a knowledge-intensive query, the target is not directly specified by the image or the query text~\citep{wang2017fvqa,marino2019ok,schwenk2022okvqa}. The query may refer to an entity through indirect factual clues, such as a role, creator, brand affiliation, release history, recent event, or relation to another entity~\citep{jiang2024mmsearch,geng2025webwatcher,wu2025mmsearch}. Solving the task therefore requires two coupled steps: first turning these clues into an explicit target hypothesis, and then mapping that hypothesis back to the image~\citep{yu2016modeling,hu2016segmentation,kamath2021mdetr}. This coupling makes the problem different from simply answering a knowledge question. Supporting clues may reveal the correct identity but provide only weak visual cues, while the image may contain multiple plausible instances, distractors, or objects with similar appearance. A model must therefore verify that the resolved identity is compatible with the observed region, rather than relying on knowledge or appearance alone. The key gap therefore lies in converting a resolved identity into a grounded visual output, not merely in recognizing the entity. 
This gap motivates Perception Deep Research for open-world visual perception, where models must actively seek external evidence, resolve the hidden identity behind a query, and ground it to concrete visual outputs.

\begin{figure}[t]
    \centering
    \includegraphics[width=1\textwidth]{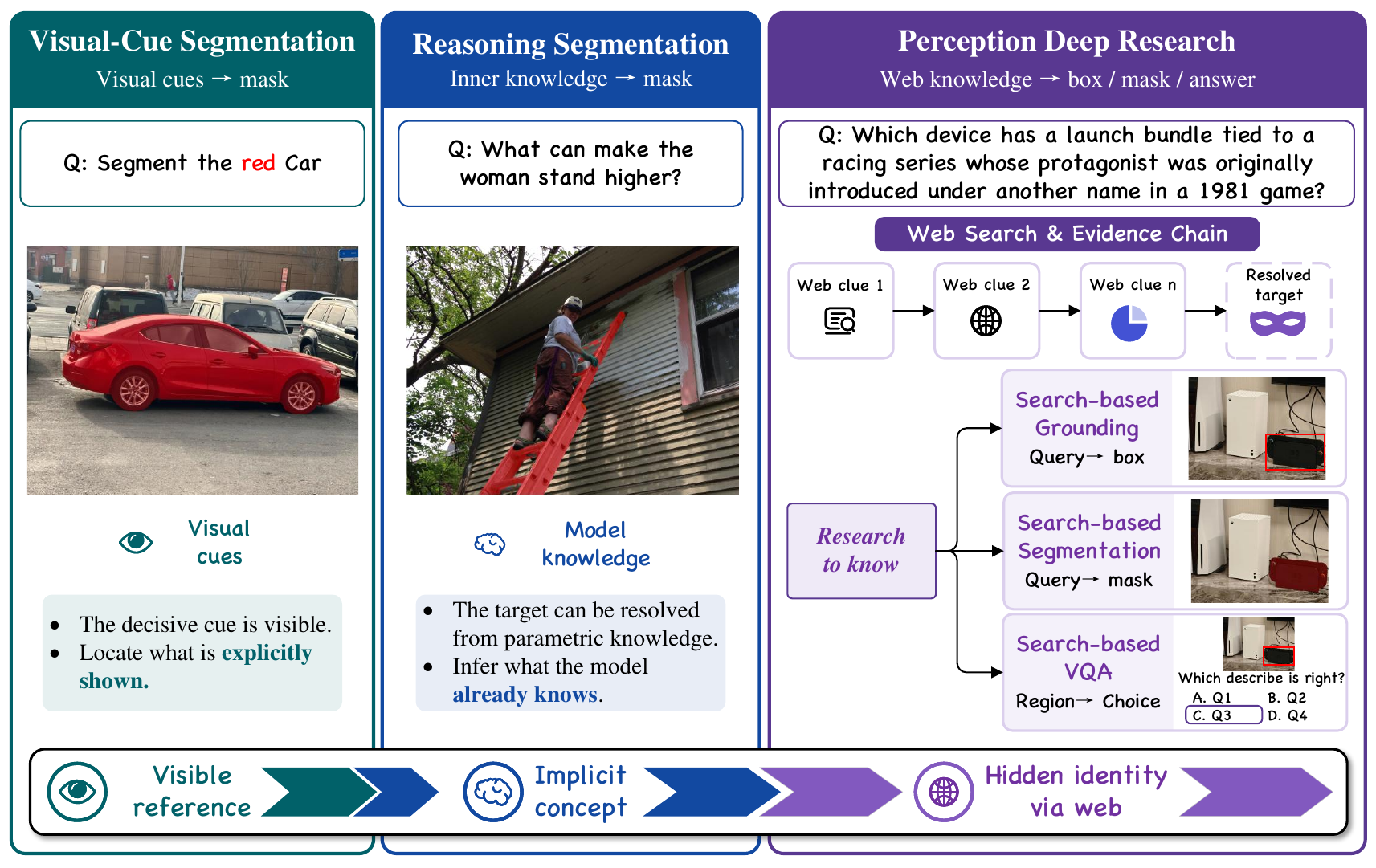}
    \vspace{-9mm}
    \caption{Our Perception Deep Research extends grounded perception from visual-cue reference and internal-knowledge reasoning to web-knowledge search.}
    \label{fig:teaser}
    \vspace{-5mm}
\end{figure}

To make Perception Deep Research measurable, we introduce \textbf{WebEyes}, an object-anchored benchmark for evidence-to-pixel visual perception. WebEyes starts from concrete visual instances and builds knowledge-intensive queries, verifiable external evidence, target identities, and spatial annotations around them. This design requires models to infer not only \textit{what} the target is, but also \textit{where} it appears. WebEyes supports three complementary task views: \textbf{Search-based Grounding} for box prediction, \textbf{Search-based Segmentation} for mask prediction, and \textbf{Search-based VQA} for region-level answer selection. Together, they evaluate whether external evidence can be reliably converted into grounded visual outputs.

\begin{figure}[t]
    \centering
    \includegraphics[width=1\linewidth]{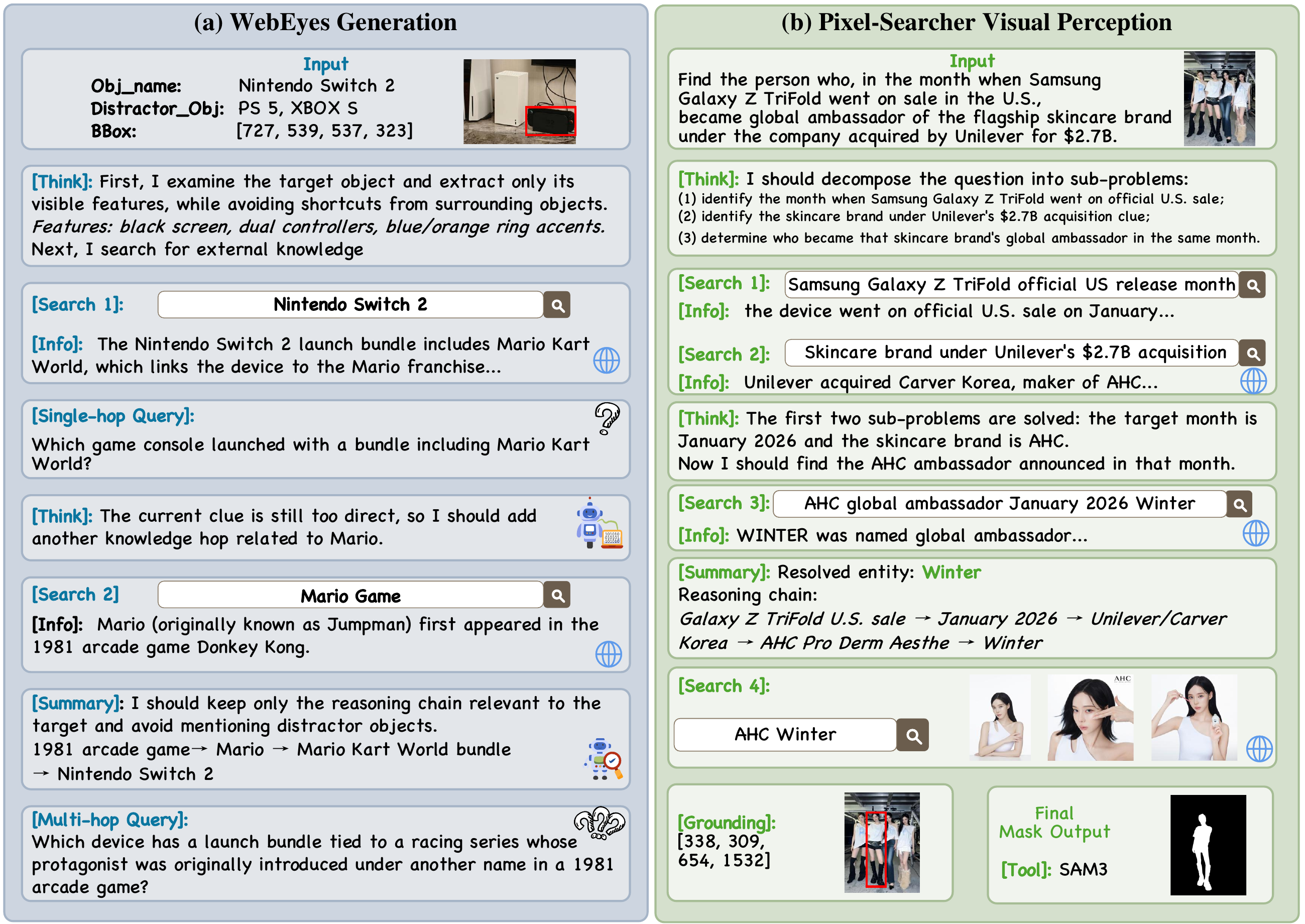}
    \vspace{-6mm}
    \caption{Overview of WebEyes generation and Pixel-Searcher inference. }
    \label{fig:compare}
    \vspace{-4mm}
\end{figure}

We further introduce \textbf{Pixel-Searcher}, an agentic search-to-pixel workflow for Perception Deep Research. It decomposes knowledge-intensive queries, gathers external evidence, resolves target identities, matches them to visual candidates, and produces the required box, mask, or answer. Experiments show that direct perception models struggle on WebEyes when decisive clues are absent from the image and frozen knowledge is insufficient. Pixel-Searcher improves performance through external evidence and step-wise reasoning, while diagnostic results show that the main bottlenecks lie in evidence acquisition, identity resolution, and visual instance binding, rather than final mask refinement.

Our contributions are threefold:
\begin{itemize}
    \item We establish \textbf{Perception Deep Research} for open-world visual perception, where models must actively seek external evidence, resolve the hidden identity behind a query, and ground it to concrete visual outputs.
    \item We construct \textbf{WebEyes}, an object-anchored benchmark with verifiable evidence, knowledge-intensive queries, precise box/mask annotations, and three task views: Search-based Grounding, Search-based Segmentation, and Search-based VQA.
    \item We propose \textbf{Pixel-Searcher}, an agentic search-to-pixel workflow, and provide diagnostic experiments that reveal key bottlenecks in evidence acquisition, identity resolution, and visual instance binding.
\end{itemize}

\section{Related Work}
\label{sec:related}

\paragraph{Visual perception with language.}
Language-guided visual perception spans referring expression comprehension, phrase grounding, and segmentation.
RefCOCO-style referring expression comprehension established a common setting where a model localizes an object from a natural-language expression and uses contextual relations among objects as key cues~\citep{yu2016modeling,wu2023onlinerefer}.
MDETR extends grounding to sentence-level phrase-region alignment by training an end-to-end detector conditioned on text spans~\citep{kamath2021mdetr}.
LISA further broadens language-guided perception by using an MLLM to interpret reasoning segmentation prompts and produce masks~\citep{lai2024lisa}.
Other grounding and segmentation methods improve open-set detection, mask prediction, video grounding, and region-level multimodal grounding~\citep{liu2024grounding,yang2022lavt,wang2022cris,shang2024prompt,wang2025adatooler,liang2025referdino,rasheed2024glamm,wu2025ragnet}.
Despite these advances, existing methods usually assume that the target can be identified from the image, the prompt, or model-internal knowledge; our work instead study cases where the target identity must first be resolved from web evidence before it can be grounded or segmented.

\paragraph{Agentic multimodal search.}
Open-knowledge VQA benchmarks such as OK-VQA expose that visual questions often require factual knowledge beyond the image~\citep{marino2019ok}.
MMSearch evaluates whether large multimodal models can act as search engines by decomposing multimodal questions, issuing searches, and synthesizing answers from retrieved evidence~\citep{jiang2024mmsearch}.
WebWatcher pushes this direction toward browsing-centric vision-language deep research agents that inspect pages and visual evidence during multi-step reasoning~\citep{geng2025webwatcher}.
Related work also studies fact-based VQA, search-augmented VQA, multimodal browsing~\citep{wang2017fvqa,feng2026gen,lin2022retrieval,yao2026mm,chen2026opensearch,tang2026rose,liang2026seg}.
However, existing work mainly studies search as an answer-synthesis tool, a browsing ability, or an auxiliary signal for segmentation.
In contrast, our work evaluates whether web evidence can be grounded to object-level outputs through shared annotations across grounding, segmentation, and VQA, while our method explicitly resolves the hidden entity before binding it to a visual instance.

\section{WebEyes Benchmark}
Perception Deep Research asks a model to find a hidden target using external evidence and connect it to a precise visual output.
Given an image and a knowledge-intensive query, the model must identify the real-world entity referred to by the query, locate the matching instance, and return a task-specific result such as a box, mask, or answer choice.
To make this setting measurable, we introduce WebEyes, a benchmark that keeps the full chain from annotated objects to web evidence, queries, and grounded targets.
We next describe its tasks, annotation pipeline, quality control, and dataset statistics.

\begin{figure}[t]
    \centering
    \includegraphics[width=1\linewidth]{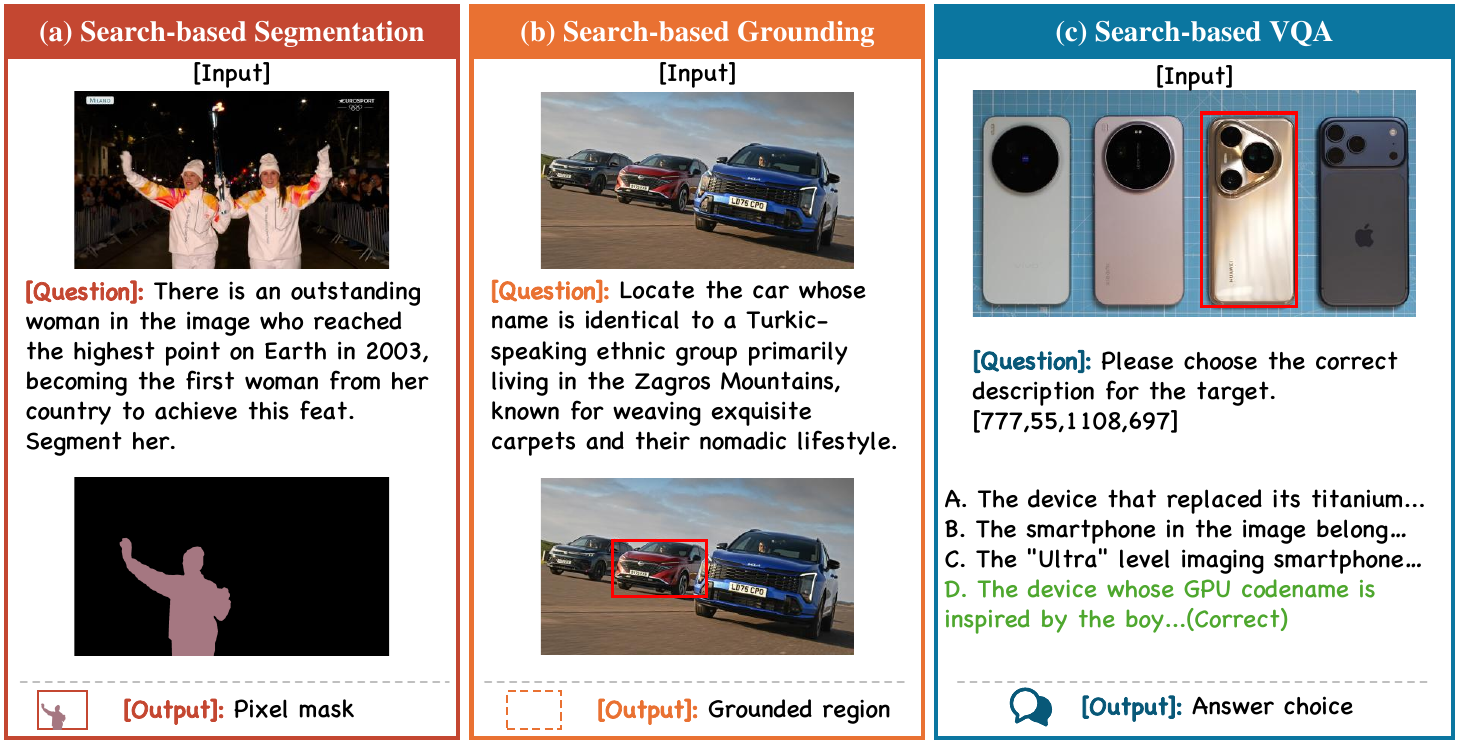}
    \vspace{-6mm}
    \caption{Examples of WebEyes task views: Search-based Segmentation outputs a mask, Search-based Grounding outputs a grounded region, and Search-based VQA selects the correct description for a highlighted target.}
    \label{fig:task_varoamts}
    \vspace{-2mm}
\end{figure}

\subsection{Benchmark Format and Statistics}

\textbf{Data format.}
As shown in Figure~\ref{fig:task_varoamts}, WebEyes supports three task views built from the same object-level annotation layer.
\emph{Search-based Grounding (SearchGround)} predicts a bounding box from the image and query, \emph{Search-based Segmentation (SearchSeg)} predicts a mask from the same input, and \emph{Search-based VQA (SearchVQA)} selects the correct knowledge-rich description for a highlighted grounded instance.
These views evaluate grounded perception from complementary perspectives: SearchGround tests whether the resolved entity can be localized, SearchSeg further measures pixel-level shape recovery, and SearchVQA checks whether a grounded region can be matched to the correct external-knowledge description.

\begin{figure}[t]
    \centering
    \begin{minipage}[c]{0.3\linewidth}
        \centering
        \includegraphics[width=\linewidth]{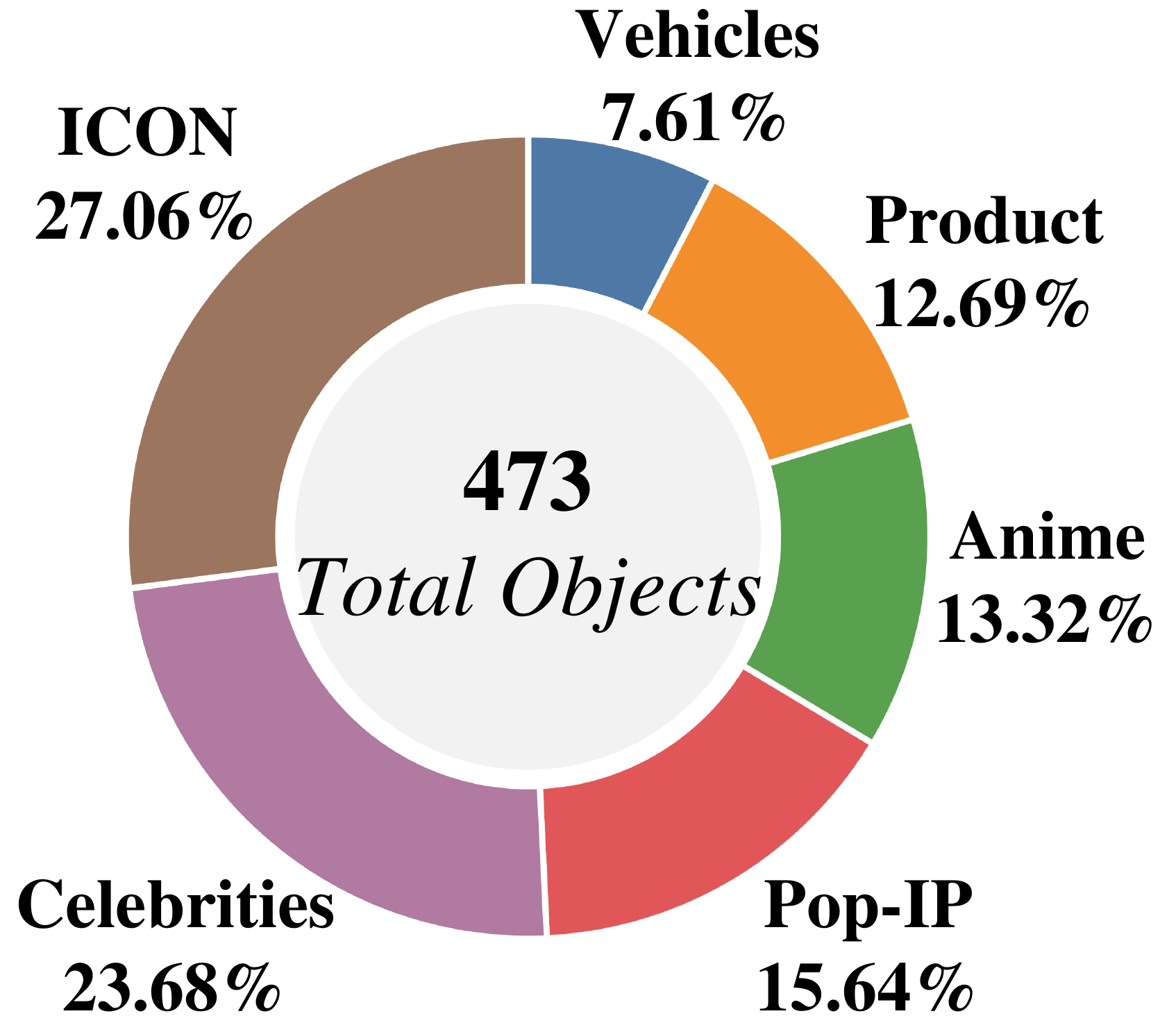}
        \caption{WebEyes Category \\ distribution.}
        \label{fig:webeyes_stats}
    \end{minipage}\hfill
    \begin{minipage}[c]{0.68\linewidth}
        \centering
        \scriptsize
        \resizebox{\linewidth}{!}{%
        \setlength\tabcolsep{2pt}
        \begin{tabular}{@{}lcccccc@{}}
        \toprule
        \multirow{2}{*}{\textbf{Benchmark}} &
        \multicolumn{2}{c}{\textbf{Knowledge Source}} &
        \multicolumn{3}{c}{\textbf{Capability}} &
        \multirow{2}{*}{\shortstack[c]{\textbf{Fine-}\\\textbf{Grained}}} \\
        \cmidrule(lr){2-3}\cmidrule(lr){4-6}
        & \textbf{Common} & \textbf{Web} & \textbf{Ground.} & \textbf{Seg.} & \textbf{QA} & \\
        \midrule
        
        RefCOCO~\citep{yu2016modeling} &
        \textcolor{red!70!black}{$\times$} &
        \textcolor{red!70!black}{$\times$} &
        \textcolor{green!60!black}{$\checkmark$} &
        \textcolor{green!60!black}{$\checkmark$} &
        \textcolor{red!70!black}{$\times$} &
        \textcolor{green!60!black}{$\checkmark$} \\
        
        ReasonSeg~\citep{lai2024lisa} &
        \textcolor{green!60!black}{$\checkmark$} &
        \textcolor{red!70!black}{$\times$} &
        \textcolor{red!70!black}{$\times$} &
        \textcolor{green!60!black}{$\checkmark$} &
        \textcolor{red!70!black}{$\times$} &
        \textcolor{green!60!black}{$\checkmark$} \\
        
        MMSearch~\citep{jiang2024mmsearch} &
        \textcolor{green!60!black}{$\checkmark$} &
        \textcolor{green!60!black}{$\checkmark$} &
        \textcolor{red!70!black}{$\times$} &
        \textcolor{red!70!black}{$\times$} &
        \textcolor{green!60!black}{$\checkmark$} &
        \textcolor{red!70!black}{$\times$} \\
        
        BrowseComp-VL~\citep{geng2025webwatcher} &
        \textcolor{green!60!black}{$\checkmark$} &
        \textcolor{green!60!black}{$\checkmark$} &
        \textcolor{red!70!black}{$\times$} &
        \textcolor{red!70!black}{$\times$} &
        \textcolor{green!60!black}{$\checkmark$} &
        \textcolor{red!70!black}{$\times$} \\
        \midrule
        
        \textbf{WebEyes} &
        \textbf{\textcolor{green!60!black}{$\checkmark$}} &
        \textbf{\textcolor{green!60!black}{$\checkmark$}} &
        \textbf{\textcolor{green!60!black}{$\checkmark$}} &
        \textbf{\textcolor{green!60!black}{$\checkmark$}} &
        \textbf{\textcolor{green!60!black}{$\checkmark$}} &
        \textbf{\textcolor{green!60!black}{$\checkmark$}} \\
        \bottomrule
        \end{tabular}%
        }
        \captionof{table}{Comparison of WebEyes with representative benchmarks.}
        \label{tab:dataset_comparison}
    \end{minipage}
    \vspace{-4mm}
\end{figure}

\textbf{Scale and categories.}
The released benchmark contains 120 images, 473 annotated object instances, and 645 unique QA pairs.
These QA pairs define 645 SearchGround samples and 645 SearchSeg samples, while 637 of them also include valid multiple-choice options for SearchVQA, yielding 1,927 task samples in total. Figure~\ref{fig:webeyes_stats} shows the category distribution, which covers a wide range of real-world entities.

\textbf{Comparison with existing benchmarks.}
As shown in Table~\ref{tab:dataset_comparison}, RefCOCO-style datasets mainly evaluate language-to-region alignment~\citep{yu2016modeling}, while ReasonSeg focuses on reasoning-based segmentation without web-based identity resolution~\citep{lai2024lisa}.
Search-oriented benchmarks such as MMSearch and BrowseComp-VL evaluate browsing ability, but their outputs are usually textual or image-level~\citep{jiang2024mmsearch,geng2025webwatcher}.
WebEyes differs by requiring the searched evidence to be grounded as box-level, mask-level, and region-level verification outputs.

\subsection{Annotation Pipeline}

\begin{figure}[t]
    \centering
    \includegraphics[width=1\textwidth]{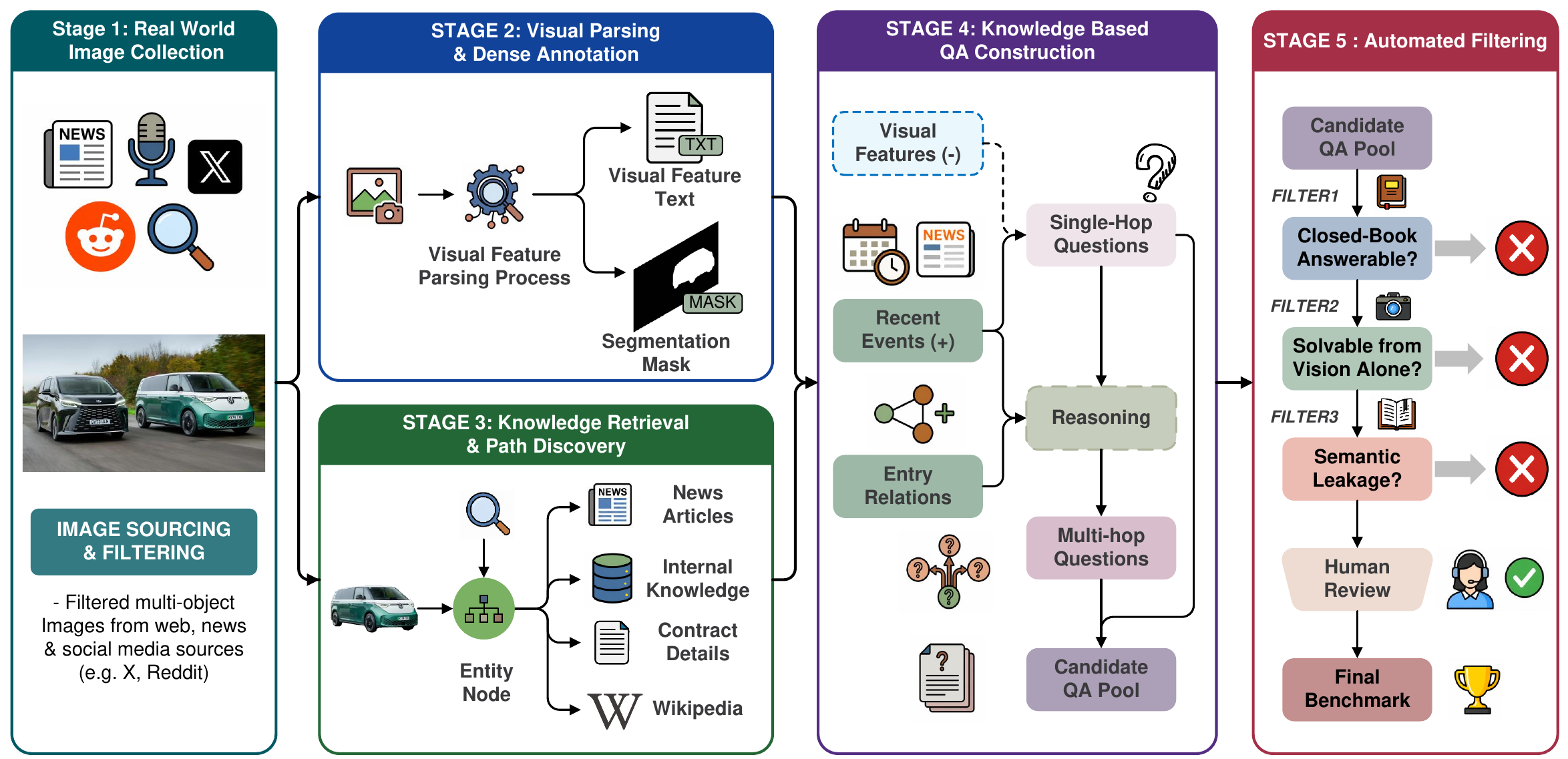}
    \vspace{-6mm}
    \caption{Automated WebEyes construction pipeline. The workflow annotates objects, links entities, searches evidence, generates questions, and filters shortcuts.}
    \label{fig:webeyes_pipeline}
    \vspace{-4mm}
\end{figure}

Figure~\ref{fig:webeyes_pipeline} shows the construction process.
WebEyes follows an object-first workflow: each annotated object is expanded into evidence paths, questions, and task instances, forming a traceable chain from mask/box supervision to external knowledge and grounded evaluation.

\textbf{Stage 1: Multi-Instance Image Collection.}
We select images primarily based on multi-instance complexity.
Candidate images are collected from web image search, news pages, and social-media posts, focusing on recent scenes involving icons, celebrities, pop-culture IPs, anime/game characters, products, and vehicles.
An MLLM-assisted screening step keeps images with multiple recognizable foreground instances and plausible distractors, while removing low-quality, text-dominated, severely occluded, or insufficiently ambiguous images.

\textbf{Stage 2: Object Annotation and Visual Parsing.}
Annotators mark foreground instances, refine masks with SAM3, and save the mask, box, object name, and category.
The agent then summarizes each instance with visual feature text describing its appearance, context, and nearby objects.
Each object therefore has visual supervision for evaluation and text cues for retrieval and question generation.

\textbf{Stage 3: Chained Evidence Retrieval and Path Discovery.}
For each annotated object, the agent performs a three-round chained search, where the result of each round conditions the next round.
The search starts by resolving the object into a searchable entity using its name, category, context, and image-checkable cues.
The resolved entity is then used with the Google Search API to retrieve public evidence within a six-month window before annotation, focusing on non-visual facts such as recent events, roles, creators, brands, product details, release histories, reports, or entity relations.
The retrieved facts are further expanded into connected evidence paths that support multi-hop questions rather than direct entity lookup.
The output is an evidence record containing the resolved entity, source URLs, access dates, visual category, and image-checkable cues.

\textbf{Stage 4: Knowledge-Based QA Construction.}
Given an evidence record, the agent generates a question by hiding the target entity name and direct visual attributes while preserving the factual clues needed to identify it.
Single-hop questions use one non-visual fact, such as a creator, brand, role, release, or recent event.
Multi-hop questions are built from the chained evidence path and require two or more facts before resolving the visible target.

\subsection{Quality Control}

Quality control combines automatic filtering of candidates solvable by shortcuts with manual verification.
The agent filters three failure modes: \textit{Closed-book shortcuts}, \textit{Vision-only shortcuts}, and \textit{Text leakage or non-uniqueness}. This step rejects 38.2\% of automatically generated candidates.

The remaining candidates enter manual review.
Human reviewers check evidence correctness, target uniqueness, text leakage, mask/box quality, and consistency across SearchGround, SearchSeg, and SearchVQA.
Among candidates that pass automatic filtering, reviewers reject another 49.2\%.
Each retained sample keeps a clear chain from source image to annotated object, external evidence, question, and grounded answer.

\begin{figure}[t]
    \centering
    \includegraphics[width=1\textwidth]{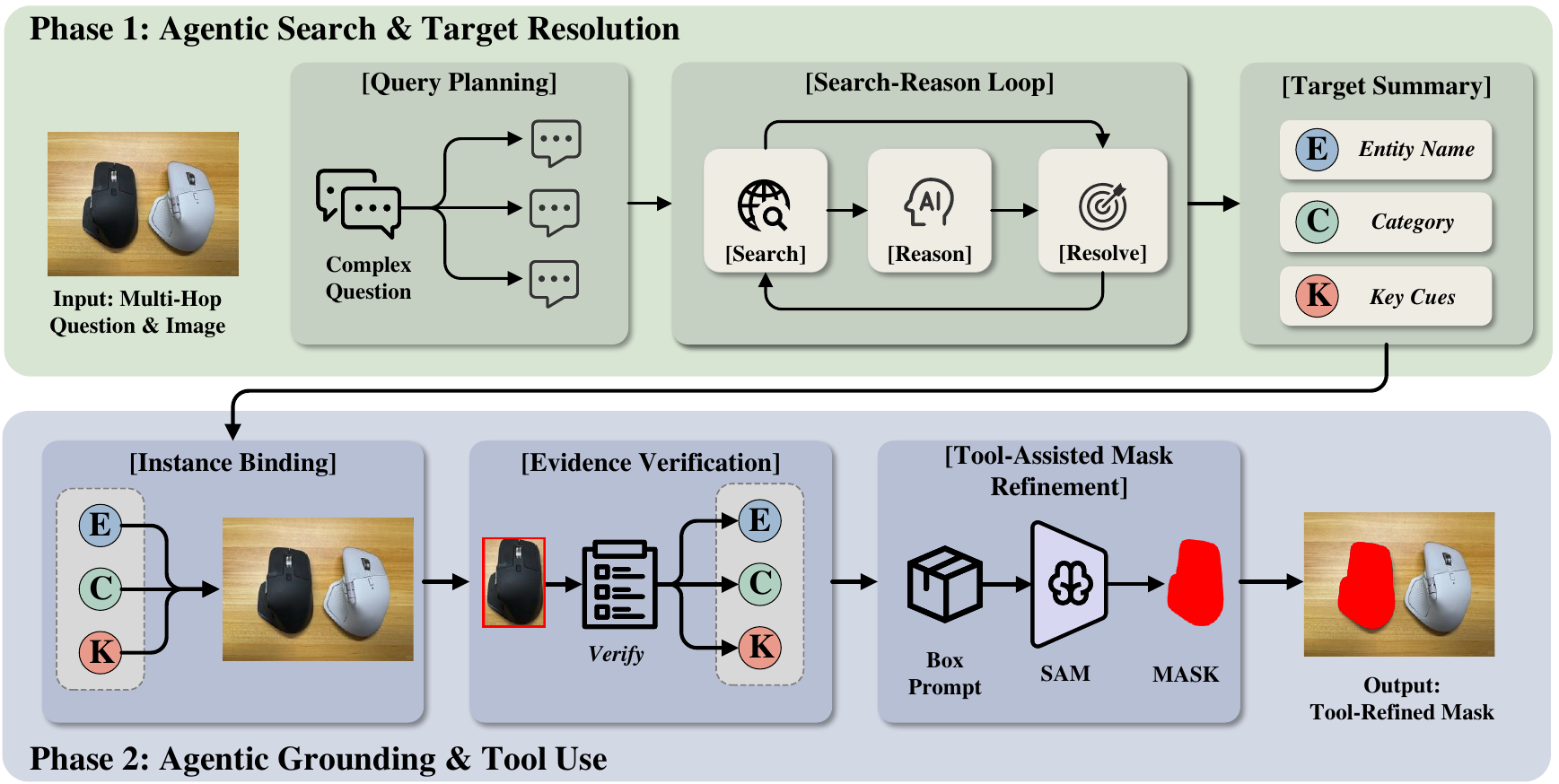}
    \vspace{-6mm}
    \caption{Pixel-Searcher overview. Forward tasks resolve the hidden entity and ground it to a box; Search-based VQA matches answer options to the highlighted region.}
    \label{fig:overall_arch}
\end{figure}

\section{Pixel-Searcher: An Agentic Search-to-Pixel Workflow}
\label{sec:methodology}

Pixel-Searcher is a reference workflow for Perception Deep Research.
Instead of treating a knowledge-intensive query as a direct grounding prompt, it converts the task into an agentic search-to-pixel process.
As shown in Figure~\ref{fig:overall_arch}, Pixel-Searcher contains two phases: \emph{Agentic Search \& Target Resolution} and \emph{Agentic Grounding \& Tool Use}.
The first phase searches for missing identity evidence and summarizes the hidden target, while the second phase binds the resolved target to a visible instance and invokes visual tools for task-specific outputs.

\subsection{Overview}

Given an image $I$ and a query $q$, Pixel-Searcher first resolves the hidden target into a structured hypothesis:
\begin{equation}
h=\{e,c,K\},
\end{equation}
where $e$ is the resolved entity name, $c$ is its visual category, and $K=\{k_j\}_{j=1}^{m}$ denotes image-checkable cues distilled from external evidence.
This hypothesis bridges web evidence and visual perception: it removes irrelevant reasoning paths from the original query and keeps the information needed for grounding, such as object type, appearance cues, identity clues, or reference evidence.

For forward tasks, Pixel-Searcher uses $h$ to bind the resolved target to a visible region in the image.
Search-based Grounding returns this verified region directly, while Search-based Segmentation further invokes a promptable segmentation tool to obtain the final mask.
For Search-based VQA, the direction is reversed: given a highlighted region, Pixel-Searcher resolves each answer option into evidence-aware cues and selects the option best supported by the grounded visual evidence.

\subsection{Agentic Search and Target Resolution}

The first phase determines what the query is actually asking the system to find.
WebEyes queries may describe targets through events, creators, brands, roles, release history, or recent news, so the target identity is often missing from the image itself.
Pixel-Searcher therefore uses an adaptive search--reason loop rather than relying on the original query alone.

The agent first plans the query and decomposes it into searchable sub-goals when needed.
It then alternates among three actions: \textsc{Search}, which retrieves external evidence; \textsc{Reason}, which connects retrieved facts and checks whether the current evidence is sufficient; and \textsc{Resolve}, which outputs the current target hypothesis.
The loop is bounded by a maximum number of rounds, but the path is adaptive: simple queries may require one factual lookup, while harder queries may require connecting multiple pieces of evidence.

Let $\mathcal{E}_{1:T}$ denote the evidence collected within at most $T$ rounds.
The resolution agent produces:
\begin{equation}
h=\mathcal{R}(q,\mathcal{E}_{1:T}).
\end{equation}
Unlike a free-form textual answer, $h$ is designed for visual grounding.
It contains the final visible entity, its coarse category, and key cues that can be checked in the image.
The agent also verifies that the resolved entity is not an intermediate clue, and repairs hypotheses that are unsupported, too generic, or inconsistent with the visual context.

\subsection{Agentic Grounding and Tool Use}

The second phase turns the resolved target hypothesis into grounded outputs.
Pixel-Searcher uses $h$ rather than the original query to guide visual grounding.
The workflow invokes grounding tools to obtain possible target regions, and then performs evidence verification to select the region most consistent with both the image and the resolved evidence:
\begin{equation}
b^{*}=\mathcal{A}_{\mathrm{bind}}(I,h).
\end{equation}
This makes grounding a tool-assisted decision process conditioned on external evidence, rather than a one-shot text-to-box prediction.

For \textbf{Search-based Grounding}, the verified region is returned as the final answer:
\begin{equation}
\hat{y}_{\mathrm{grd}}=b^{*}.
\end{equation}

For \textbf{Search-based Segmentation}, the verified region is passed to a promptable segmentation tool:
\begin{equation}
\hat{y}_{\mathrm{seg}}=\mathcal{T}_{\mathrm{seg}}(I,b^{*}),
\end{equation}
where $\mathcal{T}_{\mathrm{seg}}$ is implemented with SAM3 in our experiments.
Thus, Pixel-Searcher focuses on resolving and locating the correct instance, while the segmentation tool handles boundary refinement.

For \textbf{Search-based VQA}, the benchmark provides an image, a highlighted target region $b$, and answer options $\{o_k\}_{k=1}^{K}$.
Pixel-Searcher applies the same evidence-integration process in reverse.
It resolves each option into an entity-level summary and selects the option whose identity and visual cues best match the highlighted region:
\begin{equation}
a^{*}=\arg\max_{k}\mathcal{A}_{\mathrm{vqa}}(I,b,o_k).
\end{equation}

In this way, Pixel-Searcher provides an inspectable workflow for WebEyes.
Failures can be traced to search planning, evidence integration, target-instance binding, or tool-based mask refinement.

\section{Experiments}
\label{sec:experiments}

We evaluate whether WebEyes is challenging and Pixel-Searcher improves open-source multimodal models across grounding, segmentation, grounded answer selection, ablations, and failure analysis.

\subsection{Experimental Setup}

All methods use the same WebEyes inputs, splits, and task-specific output interfaces, without task-specific fine-tuning.
Direct baselines predict boxes from the image and query, segmentation converts the box into a mask with SAM3~\citep{kirillov2023segment}, and Search-based VQA uses the image, target box, and answer options; Pixel-Searcher differs by inserting hidden-entity search before grounding and mask refinement.
We use Qwen3-VL-8B-Instruct~\citep{qwen3technicalreport} as the general Qwen baseline because the Qwen-3.5 showed weaker instruction following in preliminary grounding trials and often failed to output valid bounding boxes.
We report percentage scores: IoU and Recall@0.5 for grounding, gIoU and cIoU for segmentation, and exact-match accuracy for Search-based VQA.

\begin{table*}[t]
    \centering
    \caption{SearchGround results. Bold marks the best result among compared open-source methods.}
    \label{tab:grounding}
    \small
    \renewcommand{\arraystretch}{1.3}
    \resizebox{\textwidth}{!}{%
        \begin{tabular}{l cccccccccccccc}
            \toprule
            \multirow{2}{*}{\textbf{Method}} & \multicolumn{2}{c}{\textbf{Vehicles}} & \multicolumn{2}{c}{\textbf{Pop-IP}} & \multicolumn{2}{c}{\textbf{Anime}} & \multicolumn{2}{c}{\textbf{ICON}} & \multicolumn{2}{c}{\textbf{Celebrities}} & \multicolumn{2}{c}{\textbf{PRODUCT}} & \multicolumn{2}{c}{\textbf{Overall}} \\
            \cmidrule(lr){2-3} \cmidrule(lr){4-5} \cmidrule(lr){6-7} \cmidrule(lr){8-9} \cmidrule(lr){10-11} \cmidrule(lr){12-13} \cmidrule(lr){14-15}
            & IoU & R@0.5 & IoU & R@0.5 & IoU & R@0.5 & IoU & R@0.5 & IoU & R@0.5 & IoU & R@0.5 & IoU & R@0.5 \\
            \midrule
            \rowcolor{rowbg} \multicolumn{15}{l}{\textit{Closed-source Models}} \\
            Doubao-Seed-1.6       & 52.16 & 63.89 & 18.07 & 20.78 & 36.69 & 37.97 & 24.26 & 36.00 & 36.37 & 39.20 & 45.77 & 55.84 & 31.53 & 38.98 \\
            Doubao-Seed-1.8       & 55.87 & 69.44 & 30.25 & 35.06 & 31.72 & 32.91 & 24.98 & 36.80 & 37.35 & 41.60 & 47.73 & 57.14 & 33.28 & 41.30 \\
            Doubao-Seed-2.0-Pro   & 60.73 & 75.00 & 29.04 & 32.47 & 36.32 & 39.24 & 27.37 & 40.00 & 38.32 & 43.20 & 52.76 & 63.64 & 35.69 & 44.41 \\
            GPT-4.1               & 30.01 & 27.78 & 3.95 & 1.30 & 6.59 & 6.33 & 1.61 & 0.00 & 6.29 & 2.40 & 13.15 & 7.79 & 6.38 & 3.88 \\
            GPT-5.4               & 34.36 & 41.67 & 19.64 & 22.08 & 13.90 & 7.59 & 18.89 & 19.60 & 20.08 & 17.60 & 19.46 & 18.18 & 19.53 & 19.10 \\
            GPT-4o                & 13.94 & 2.78 & 1.14 & 0.00 & 1.09 & 0.00 & 1.47 & 0.00 & 0.17 & 0.00 & 6.59 & 2.60 & 2.44 & 0.47 \\
            Gemini-3.1-Flash-Lite & 53.57 & 63.89 & 26.05 & 23.38 & 30.94 & 22.78 & 15.08 & 18.00 & 37.71 & 37.60 & 40.33 & 41.56 & 27.90 & 28.42 \\
            Gemini-3.1-Pro        & 62.89 & 75.00 & 32.13 & 33.77 & 24.14 & 22.78 & 22.22 & 29.60 & 31.64 & 32.00 & 45.43 & 53.25 & 30.52 & 35.09 \\
            \midrule
            \rowcolor{rowbg} \multicolumn{15}{l}{\textit{Open-Source Grounding Models}} \\
            Perception-R1~\citep{yu2026perception}         & 16.07 & 11.11 & 9.61 & 6.58 & 22.16 & 21.52 & 4.08 & 4.00 & 23.50 & 22.40 & 13.15 & 12.99 & 14.76 & 13.10 \\
            UniVG-R1~\citep{bai2025univg}              & 29.29 & 33.33 & 19.02 & 22.37 & 31.45 & 31.65 & 14.63 & 19.92 & 34.68 & 37.60 & 25.67 & 28.57 & 25.79 & 28.91 \\
            Ground-R1~\citep{cao2025ground}             & 44.47 & 50.00 & 24.11 & 28.95 & 24.23 & 24.05 & 9.60 & 13.67 & 32.41 & 32.80 & 25.23 & 28.57 & 26.68 & 29.67 \\
            \midrule
            \rowcolor{rowbg} \multicolumn{15}{l}{\textit{Open-Source General Models}} \\
            OneThinker-8B~\citep{feng2025onethinker}         & 51.19 & \textbf{61.11} & \textbf{28.19} & \textbf{32.89} & 23.86 & 24.05 & 21.81 & 32.81 & \textbf{37.71} & \textbf{42.40} & 33.92 & 38.96 & 32.78 & 38.70 \\
            InternVL-3.5-8B~\citep{wang2025internvl3}       & 10.58 & 2.78 & 3.49 & 0.00 & 3.39 & 1.27 & 1.45 & 0.00 & 5.92 & 1.60 & 7.75 & 1.30 & 4.07 & 0.78 \\
            Qwen3-VL-8B~\citep{qwen3technicalreport}              & 47.58 & 58.33 & 20.13 & 25.97 & 24.78 & 29.11 & 22.30 & 28.40 & 32.54 & 36.80 & 31.17 & 37.66 & 26.81 & 32.61 \\
            \midrule
            \rowcolor{rowbg} \multicolumn{15}{l}{\textit{Proposed Method}} \\
            Pixel-Searcher (Ours) & \textbf{54.74} & 58.33 & 23.53 & 24.68 & \textbf{38.81} & \textbf{43.04} & \textbf{29.68} & \textbf{42.80} & 35.94 & 38.40 & \textbf{42.12} & \textbf{48.05} & \textbf{34.17} & \textbf{41.30} \\
            \bottomrule
        \end{tabular}
    }
\end{table*}

\subsection{Main Results on WebEyes}

The main results show that WebEyes remains challenging and that resolving the hidden entity before visual prediction improves open-source models across all three task views.

\textbf{Search-based Grounding.}
Table~\ref{tab:grounding} reports Search-based Grounding results.
Pixel-Searcher is the strongest open-source method, improving Qwen3-VL-8B from 26.81 to 34.17 IoU and from 32.61 to 41.30 R@0.5.
The gains are clearest in ambiguity-heavy categories such as \emph{Anime} and \emph{ICON}, although translating external evidence into precise boxes remains difficult.


\begin{center}
\begin{minipage}{\textwidth}
    \centering
    \captionsetup{hypcap=false}
    \captionof{table}{SearchSeg results. Bold marks the best result among compared open-source methods.}
    \label{tab:segmentation}
    \small
    \renewcommand{\arraystretch}{1.2}
    \resizebox{\textwidth}{!}{%
        \begin{tabular}{l cccccccccccccc}
            \toprule
            \multirow{2}{*}{\textbf{Method}} & \multicolumn{2}{c}{\textbf{Vehicles}} & \multicolumn{2}{c}{\textbf{Pop-IP}} & \multicolumn{2}{c}{\textbf{Anime}} & \multicolumn{2}{c}{\textbf{ICON}} & \multicolumn{2}{c}{\textbf{Celebrities}} & \multicolumn{2}{c}{\textbf{PRODUCT}} & \multicolumn{2}{c}{\textbf{Overall}} \\
            \cmidrule(lr){2-3} \cmidrule(lr){4-5} \cmidrule(lr){6-7} \cmidrule(lr){8-9} \cmidrule(lr){10-11} \cmidrule(lr){12-13} \cmidrule(lr){14-15}
            & gIoU & cIoU & gIoU & cIoU & gIoU & cIoU & gIoU & cIoU & gIoU & cIoU & gIoU & cIoU & gIoU & cIoU \\
            \midrule
            \rowcolor{rowbg} \multicolumn{15}{l}{\textit{Closed-source Models}} \\
            Doubao-Seed-1.6       & 65.92 & 57.13 & 23.72 & 20.66 & 40.88 & 32.60 & 63.81 & 49.15 & 47.01 & 30.39 & 54.44 & 44.05 & 50.38 & 36.13 \\
            Doubao-Seed-1.8       & 71.40 & 62.57 & 38.10 & 28.45 & 42.05 & 23.02 & 62.67 & 47.76 & 46.66 & 31.20 & 55.45 & 44.42 & 52.64 & 36.06 \\
            Doubao-Seed-2.0-Pro   & 79.00 & 71.81 & 41.66 & 21.22 & 49.79 & 36.87 & 73.26 & 61.05 & 56.21 & 36.78 & 62.13 & 51.53 & 61.22 & 43.32 \\
            GPT-4.1               & 24.28 & 20.34 & 0.85 & 0.49 & 8.10 & 6.98 & 0.43 & 0.24 & 5.84 & 3.61 & 7.19 & 4.93 & 4.99 & 6.79 \\
            GPT-5.4               & 43.03 & 37.06 & 20.90 & 10.21 & 9.74 & 7.43 & 39.43 & 29.94 & 16.48 & 8.64 & 18.93 & 14.55 & 25.09 & 13.54 \\
            GPT-4o                & 2.18 & 2.81 & 0.45 & 0.44 & 6.23 & 5.14 & 1.25 & 0.70 & 0.14 & 0.08 & 13.65 & 8.47 & 3.42 & 4.85 \\
            Gemini-3.1-Flash-Lite & 67.32 & 65.00 & 28.44 & 24.87 & 29.11 & 25.07 & 40.43 & 26.17 & 38.40 & 33.29 & 45.04 & 43.51 & 39.11 & 35.58 \\
            Gemini-3.1-Pro        & 79.79 & 72.27 & 43.80 & 27.46 & 38.14 & 26.75 & 61.19 & 48.89 & 43.89 & 28.26 & 65.06 & 49.86 & 54.56 & 38.76 \\
            \midrule
            \rowcolor{rowbg} \multicolumn{15}{l}{\textit{Open-Source Segmentation Models}} \\
            LISA-7B~\citep{lai2024lisa}               & 15.74 & 20.84 & 11.85 & 18.78 & 9.09 & 5.38 & 2.82 & 4.32 & 21.02 & 16.97 & 8.67 & 10.65 & 11.53 & 12.82 \\
            Seg-Zero-7B~\citep{liu2025seg}           & 49.43 & 35.98 & 32.54 & 16.17 & 25.79 & 21.18 & 29.57 & 18.66 & \textbf{39.26} & 27.36 & 27.58 & 15.70 & 34.03 & 22.51 \\
            Seg-R1-7B~\citep{you2025seg}             & 56.54 & 45.06 & 26.72 & 18.93 & 17.57 & 14.78 & 27.89 & 17.10 & 36.32 & 24.75 & 27.21 & 21.68 & 32.04 & 23.72 \\
            Affordance-R1~\citep{wang2026affordance}         & 11.84 & 9.49 & 13.28 & 6.93 & 14.15 & 14.28 & 30.08 & 19.27 & 9.87 & 7.98 & 15.82 & 11.02 & 15.84 & 11.50 \\
            SAM3-Agent~\citep{carion2025sam3segmentconcepts}            & 38.22 & 37.35 & 14.51 & \textbf{24.98} & 13.76 & 23.58 & 12.14 & 9.64 & 19.88 & 26.54 & 20.49 & 24.94 & 19.83 & 24.50 \\
            \midrule
            \rowcolor{rowbg} \multicolumn{15}{l}{\textit{Open-Source General Models}} \\
            OneThinker-8B~\citep{feng2025onethinker}         & 48.35 & 35.41 & 27.87 & 17.64 & 32.32 & 24.52 & 31.66 & 20.57 & 38.37 & 23.20 & 35.01 & 25.43 & 35.60 & 24.46 \\
            InternVL-3.5-8B~\citep{wang2025internvl3}       & 10.62 & 10.45 & 3.72 & 2.31 & 4.59 & 3.67 & 2.99 & 2.06 & 7.67 & 4.95 & 5.53 & 5.82 & 5.28 & 5.28 \\
            Qwen3-VL-8B~\citep{qwen3technicalreport}                & 57.63 & 46.43 & \textbf{33.82} & 11.23 & 32.12 & 26.28 & 32.84 & \textbf{20.85} & 36.81 & 24.70 & 37.43 & 25.59 & 35.78 & 25.94 \\
            \midrule
            \rowcolor{rowbg} \multicolumn{15}{l}{\textit{Proposed Method}} \\
            Pixel-Searcher (Ours) & \textbf{61.01} & \textbf{53.19} & 23.91 & 15.25 & \textbf{37.57} & \textbf{33.21} & \textbf{39.86} & 19.19 & 37.57 & \textbf{33.62} & \textbf{46.33} & \textbf{35.24} & \textbf{39.17} & \textbf{32.41} \\
            \bottomrule
        \end{tabular}
    }
    \vspace{-2mm}

    \centering
    \captionof{table}{SearchVQA results. Bold marks the best accuracy among compared open-source methods.}
    \label{tab:multichoice}
    \scriptsize
    \renewcommand{\arraystretch}{1.12}
    \setlength{\tabcolsep}{3pt}
    \resizebox{\textwidth}{!}{
        \begin{tabular}{p{6cm} ccccccc}
            \toprule
            \textbf{Method} & \textbf{Vehicles} & \textbf{Pop-IP} & \textbf{Anime} & \textbf{ICON} & \textbf{Celebrities} & \textbf{PRODUCT} & \textbf{Overall} \\
            \midrule
            \rowcolor{rowbg} \multicolumn{8}{l}{\textit{Closed-source Models}} \\
            Doubao-Seed-1.6       & 63.89 & 29.87 & 41.77 & 66.80 & 53.60 & 53.25 & 54.97 \\
            Doubao-Seed-1.8       & 69.44 & 28.57 & 41.77 & 78.80 & 54.40 & 64.94 & 61.34 \\
            Doubao-Seed-2.0-Pro   & 69.44 & 42.86 & 45.57 & 81.20 & 56.80 & 68.83 & 65.37 \\
            GPT-4.1               & 58.33 & 24.68 & 37.97 & 40.00 & 39.20 & 51.95 & 40.22 \\
            GPT-5.4               & 44.44 & 28.57 & 35.44 & 60.00 & 48.80 & 53.25 & 49.38 \\
            GPT-4o                & 58.33 & 24.68 & 27.85 & 20.40 & 39.20 & 40.26 & 29.97 \\
            Gemini-3.1-Flash-Lite & 55.56 & 23.38 & 43.04 & 40.40 & 40.80 & 49.35 & 40.68 \\
            Gemini-3.1-Pro        & 83.33 & 48.05 & 44.30 & 68.00 & 62.40 & 79.22 & 63.82 \\
            \midrule
            \rowcolor{rowbg} \multicolumn{8}{l}{\textit{Open-Source QA Models}} \\
            UniVG-R1~\citep{bai2025univg}              & 30.56 & 19.74 & 31.65 & 22.13 & 36.00 & 22.08 & 26.22 \\
            SophiaVL-R1~\citep{fan2025sophiavl}           & 38.89 & 23.68 & 22.78 & 27.46 & 34.40 & 37.66 & 29.67 \\
            VL-Rethinker~\citep{wang2025vl}          & 33.33 & 23.68 & 31.65 & 27.05 & 33.60 & 29.87 & 29.20 \\
            Vision-R1~\citep{huang2025vision}             & 13.89 & 22.37 & 17.72 & 27.87 & 12.80 & 19.48 & 21.19 \\
            \midrule
            \rowcolor{rowbg} \multicolumn{8}{l}{\textit{Open-Source General Models}} \\
            OneThinker-8B~\citep{feng2025onethinker}            & 36.11 & 21.05 & 29.11 & 24.59 & 40.00 & 23.38 & 28.26 \\
            InternVL-3.5-8B~\citep{wang2025internvl3}       & 25.00 & \textbf{29.87} & 32.91 & 42.00 & 36.00 & 31.17 & 36.02 \\
            Qwen3-VL-8B~\citep{qwen3technicalreport}               & \textbf{47.22} & 24.68 & 31.65 & 35.60 & \textbf{43.20} & 38.96 & 36.34 \\
            \midrule
            \rowcolor{rowbg} \multicolumn{8}{l}{\textit{Proposed Method}} \\
            Pixel-Searcher (Ours) & 38.89 & 24.68 & \textbf{34.18} & \textbf{50.40} & 42.40 & \textbf{42.86} & \textbf{42.24} \\
            \bottomrule
        \end{tabular}
    }
    \vspace{-1mm}
\end{minipage}
\end{center}

\textbf{Search-based Segmentation.}
Table~\ref{tab:segmentation} reports Search-based Segmentation results.
Pixel-Searcher again ranks first among open-source methods, improving Qwen3-VL-8B from 35.78 to 39.17 gIoU and from 25.94 to 32.41 cIoU.
Category-level gains are strongest in \emph{Vehicles}, \emph{Anime}, and \emph{Product}, indicating that better hidden-entity grounding transfers to box-prompted SAM3 refinement.

\textbf{Search-based VQA.}
Table~\ref{tab:multichoice} reports Search-based VQA accuracy.
Pixel-Searcher improves Qwen3-VL-8B from 36.34 to 42.24 overall accuracy and performs best among open-source methods, with clear gains in \emph{Icons} and \emph{Product}; the smaller margin to closed-source models suggests that fine-grained semantic comparison also matters.

These gains are consistent with the benchmark design, where many samples require selecting one instance among several similar objects. WebEyes does not only ask whether a model can segment or localize, but whether it can first recover the hidden target identity from external evidence. In these cases, the key decision is often instance-level verification: the model must reject visually plausible regions whose identity is inconsistent with the retrieved evidence.

The remaining gap to closed-source systems indicates that search-conditioned perception is still limited by evidence selection, entity resolution, and matching the entity to the right image region rather than by a single output format. In many samples, several plausible objects are visible, and the decisive clue only emerges after external evidence resolution, unlike standard referring perception where visual attributes usually identify the target directly. This makes errors in the search stage especially costly, since an incorrect or unclear entity can still lead to a visually reasonable but semantically wrong region.

\subsection{Ablation and Failure Analysis}
\label{subsec:ablation}

\begin{wraptable}{r}{0.5\textwidth}
    \centering
    \vspace{-2.2em}
    \caption{Component ablation of Pixel-Searcher.}
    \label{tab:ablation}
    \footnotesize
    \setlength{\tabcolsep}{2.5pt}
    \begin{tabular}{lcccc}
        \toprule
        Variant  & IoU & R@0.5 & gIoU & cIoU \\
        \midrule
        \rowcolor{rowbg} Full  & 34.17 & 41.30 & 39.17 & 32.41 \\
        w/o contradiction  & 31.34 & 37.58 & 36.51 & 28.41 \\
        w/o direct bonus  & 30.80 & 36.96 & 35.67 & 27.92 \\
        support only  & 29.84 & 35.40 & 34.64 & 27.68 \\
        w/o ref. match  & 29.00 & 34.01 & 33.05 & 26.47 \\
        direct only  & 22.28 & 27.48 & 26.49 & 24.48 \\
        w/o direct cand.  & 20.14 & 19.72 & 20.14 & 15.71 \\
        \bottomrule
    \end{tabular}
    \vspace{-1.2em}
\end{wraptable}

The ablation study asks which parts of Pixel-Searcher's evidence-to-region process are responsible for the final gains. Table~\ref{tab:ablation} removes or simplifies individual grounding cues while measuring both box quality and downstream mask quality, since Search-based Segmentation depends heavily on whether the resolved entity is first mapped to the correct instance.

The largest drops come from removing direct localization cues. Without direct candidates, IoU falls from 34.17 to 20.14 and R@0.5 falls from 41.30 to 19.72, while gIoU/cIoU drop from 39.17/32.41 to 20.14/15.71. However, the direct-only variant is also much weaker than the full system, reaching only 22.28 IoU and 26.49 gIoU. Reference matching and contradiction checking add smaller but consistent gains, showing that direct grounding must be combined with resolved entity evidence and visual verification.

Most failed segmentation samples are search/entity errors: among 389 failures, 304 are search/entity errors, 75 are entity-correct region errors, and only 10 are box-to-mask transfer errors. Thus, about three quarters of the errors occur before the system selects the correct entity, while very few come from converting a localized box into a mask. The remaining entity-correct region errors show that visual matching is still nontrivial even after the external evidence is correct. Overall, WebEyes stresses search planning, entity disambiguation, and visual instance verification under realistic distractors more than mask generation alone.

\section{Conclusion}
\label{sec:conclusion}

This paper introduces perception deep research, where web-derived evidence must be converted into box-level and pixel-level predictions.
We build WebEyes and instantiate the setting with Pixel-Searcher, showing that agentic search consistently improves Search-based Grounding, Search-based Segmentation, and Search-based VQA when visual appearance alone is insufficient.
The failure analysis shows that the dominant bottleneck is not mask refinement after a correct box, but the earlier path from search planning to entity resolution and visual instance verification.
WebEyes provides benchmark infrastructure for this direction, and Pixel-Searcher offers a simple starting workflow for studying how agentic search can identify the right entity and bind it to the right visual instance.


{
    \small
    \bibliographystyle{unsrt}
    \bibliography{reference}
}

\clearpage

\newpage
\appendix

\newpage

\section{Dataset Samples}
\label{app:samples}

Table~\ref{tab:appendix_samples} shows five representative WebEyes source images.
For each selected image, all annotated objects are listed with their overlaid target masks, object names, and corresponding knowledge-intensive queries.

\begingroup
\small
\setlength{\tabcolsep}{3pt}
\renewcommand{\arraystretch}{1.08}
\newcommand{\sourceimg}[1]{\includegraphics[width=\linewidth,height=0.115\textheight,keepaspectratio]{#1}}
\newcommand{\sourcecell}[2]{%
    \begin{minipage}[c][#1][c]{\linewidth}
        \centering
        \sourceimg{#2}
    \end{minipage}}
\newcommand{\maskimg}[1]{\includegraphics[width=\linewidth,height=0.092\textheight,keepaspectratio]{#1}}
\newcommand{\sampleheader}{%
    \begin{tabular}{@{}>{\centering\arraybackslash}m{0.19\textwidth}
                    >{\raggedright\arraybackslash}m{0.14\textwidth}
                    >{\raggedright\arraybackslash}m{0.39\textwidth}@{}}
        \textbf{Mask} & \textbf{Object} & \textbf{Query}
    \end{tabular}}
\newcommand{\sampleobjects}[1]{%
    \begin{tabular}{@{}>{\centering\arraybackslash}m{0.19\textwidth}
                    >{\raggedright\arraybackslash}m{0.14\textwidth}
                    >{\raggedright\arraybackslash}m{0.39\textwidth}@{}}
        #1
    \end{tabular}}
\begin{longtable}{>{\centering\arraybackslash}m{0.19\textwidth}
                  >{\centering\arraybackslash}m{0.77\textwidth}}
    \caption{Five WebEyes source images with all annotated objects, overlaid target masks, object names, and queries.}
    \label{tab:appendix_samples}\\
    \toprule
    \textbf{Image} & \sampleheader \\
    \midrule
    \endfirsthead
    \toprule
    \textbf{Image} & \sampleheader \\
    \midrule
    \endhead
    \midrule
    \multicolumn{2}{r}{Continued on next page} \\
    \endfoot
    \bottomrule
    \endlastfoot

    \sourcecell{0.285\textheight}{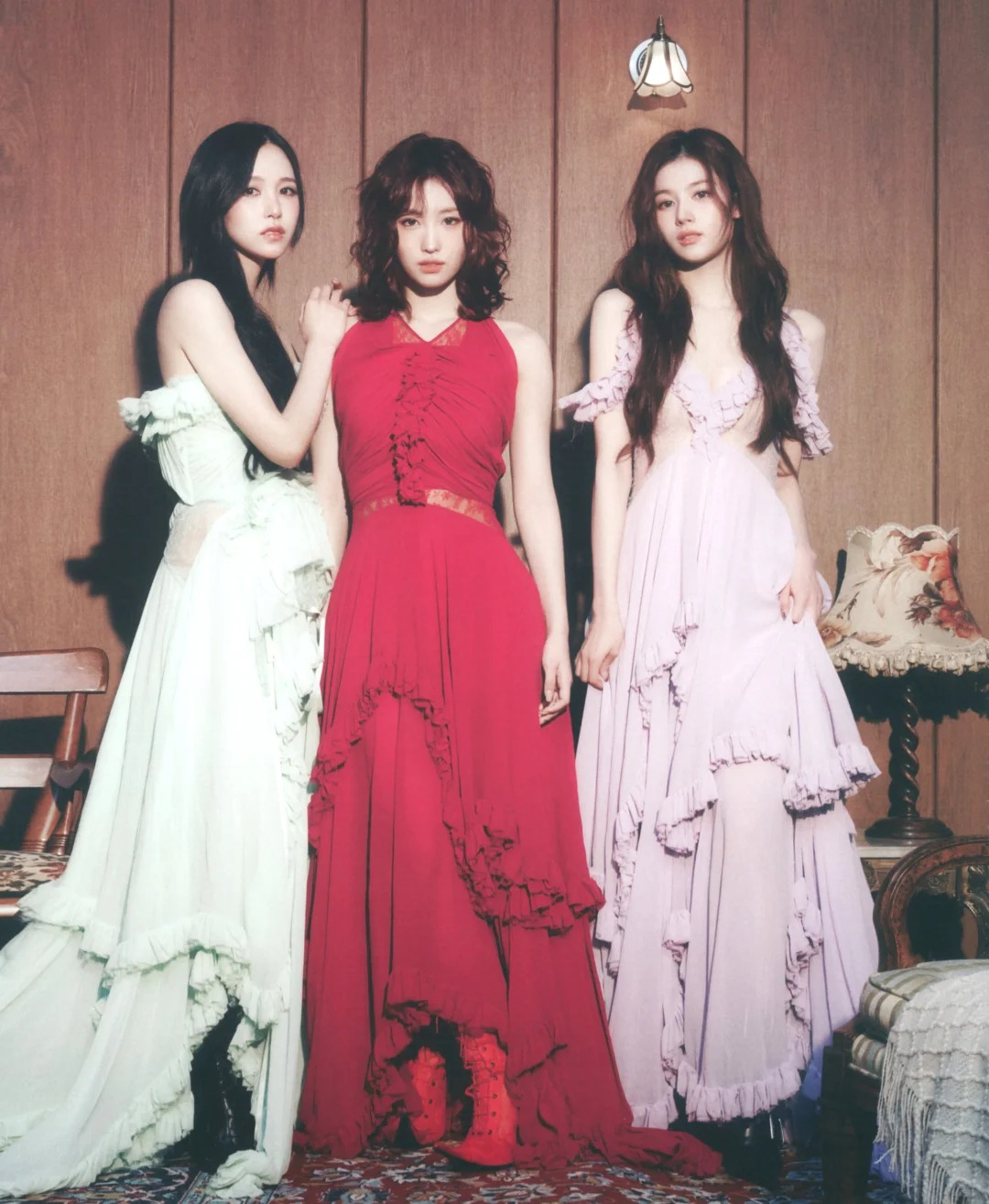} &
    \sampleobjects{
        \maskimg{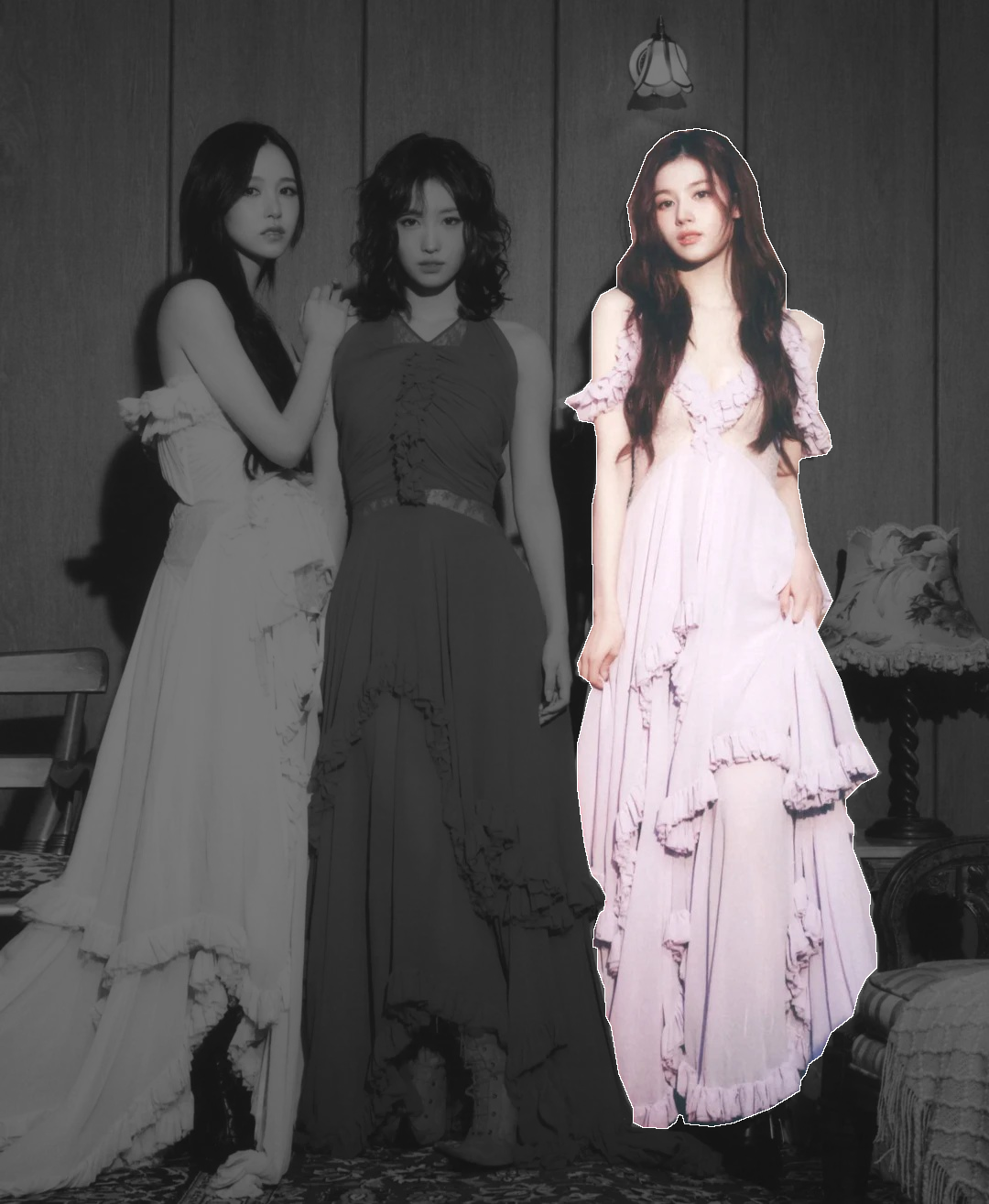} &
        Sana &
        Please find the person who became a brand ambassador for the South Korean brand NE:AR in 2025 in the image. \\
        \cmidrule(l){1-3}
        \maskimg{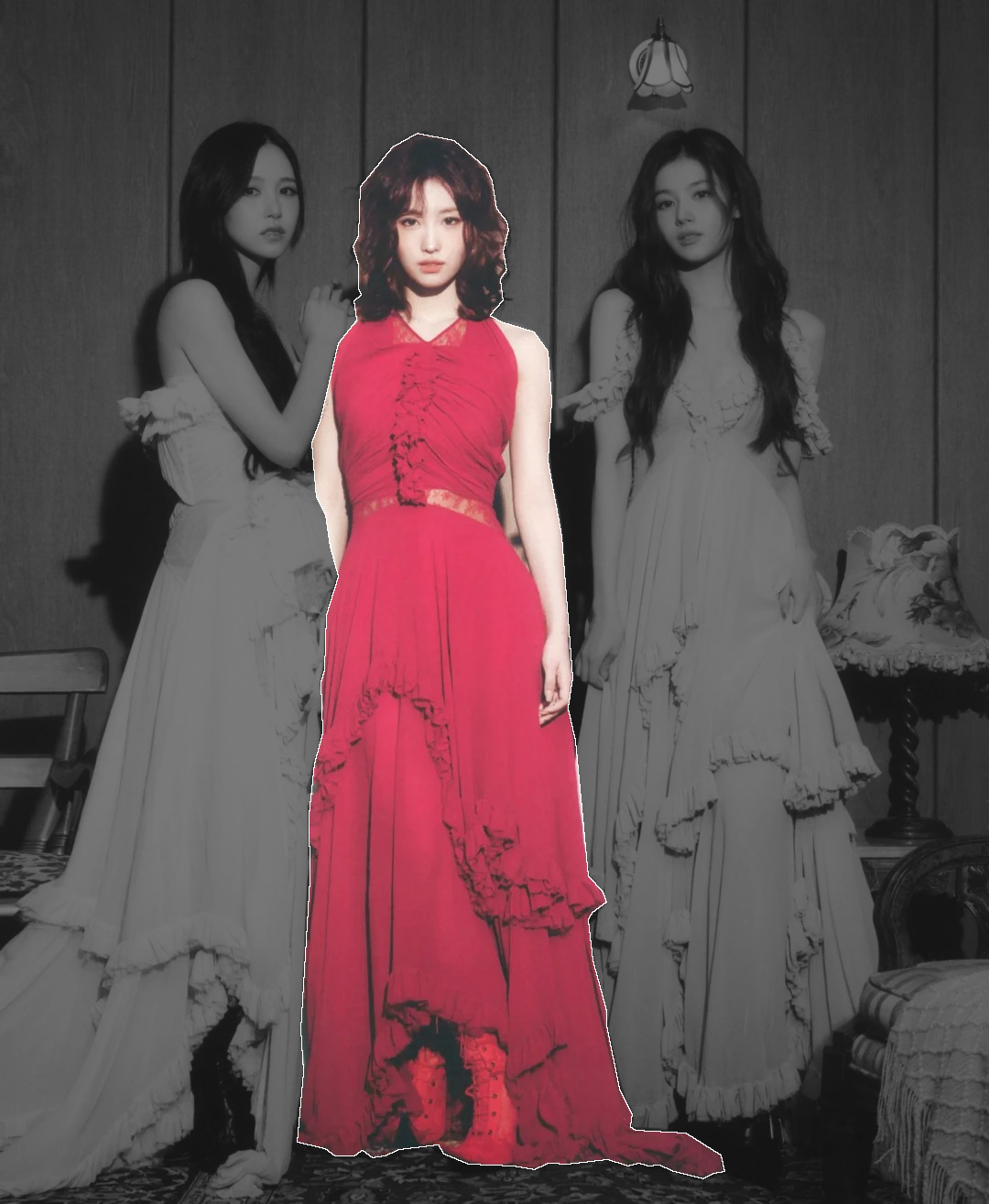} &
        Momo &
        Please find the person who performed on stage as a member of a K-pop girl group at the 2025 fashion show held by the brand that collaborated with the film The Bride! to launch a limited four-piece capsule collection in the image. \\
        \cmidrule(l){1-3}
        \maskimg{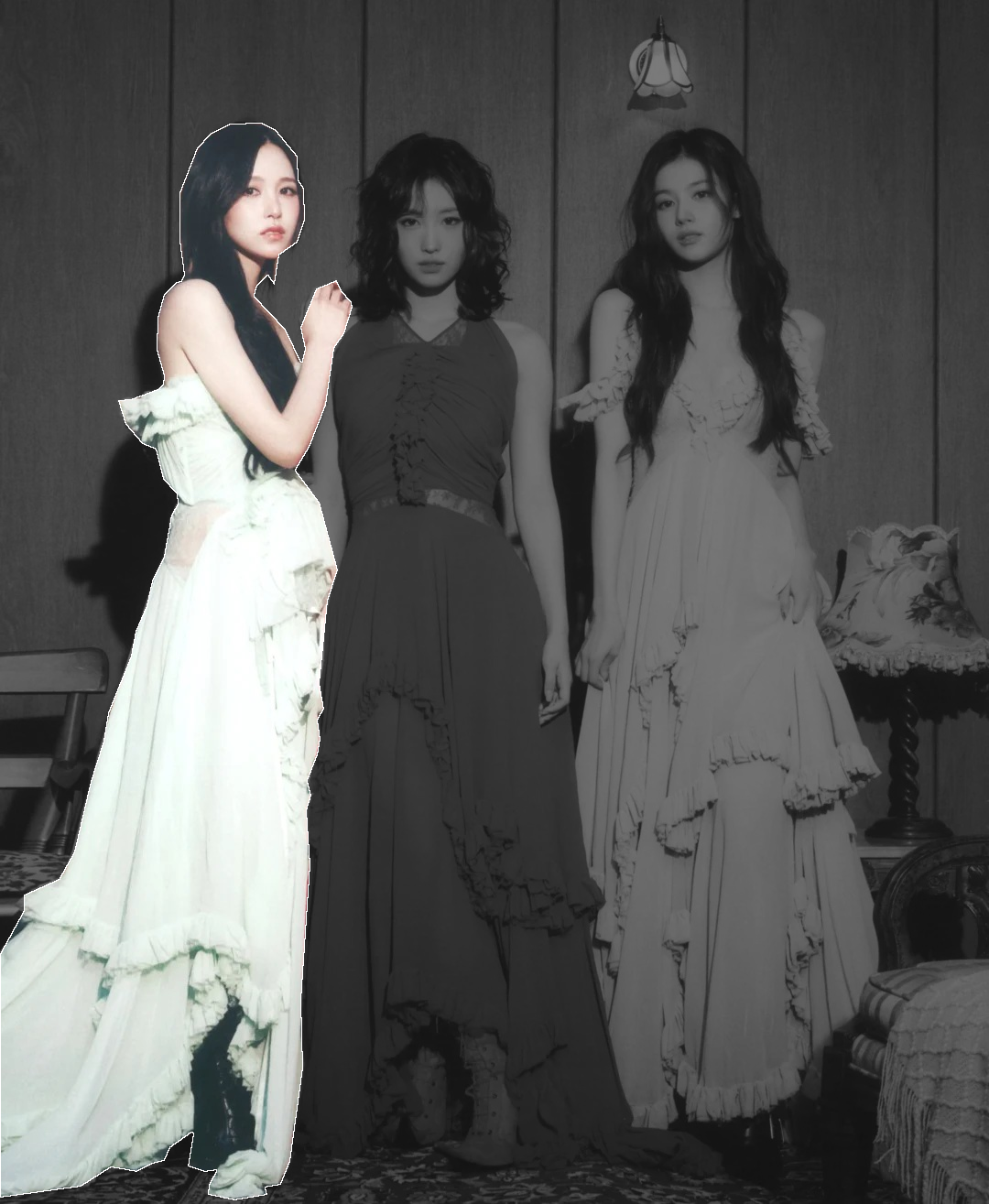} &
        Mina &
        Please find the person who endorses the brand whose name simultaneously encodes the five values of TOUCH, TECH, TEMPO, TREND, and TIME in the image.
    } \\
    \midrule

    \sourcecell{0.180\textheight}{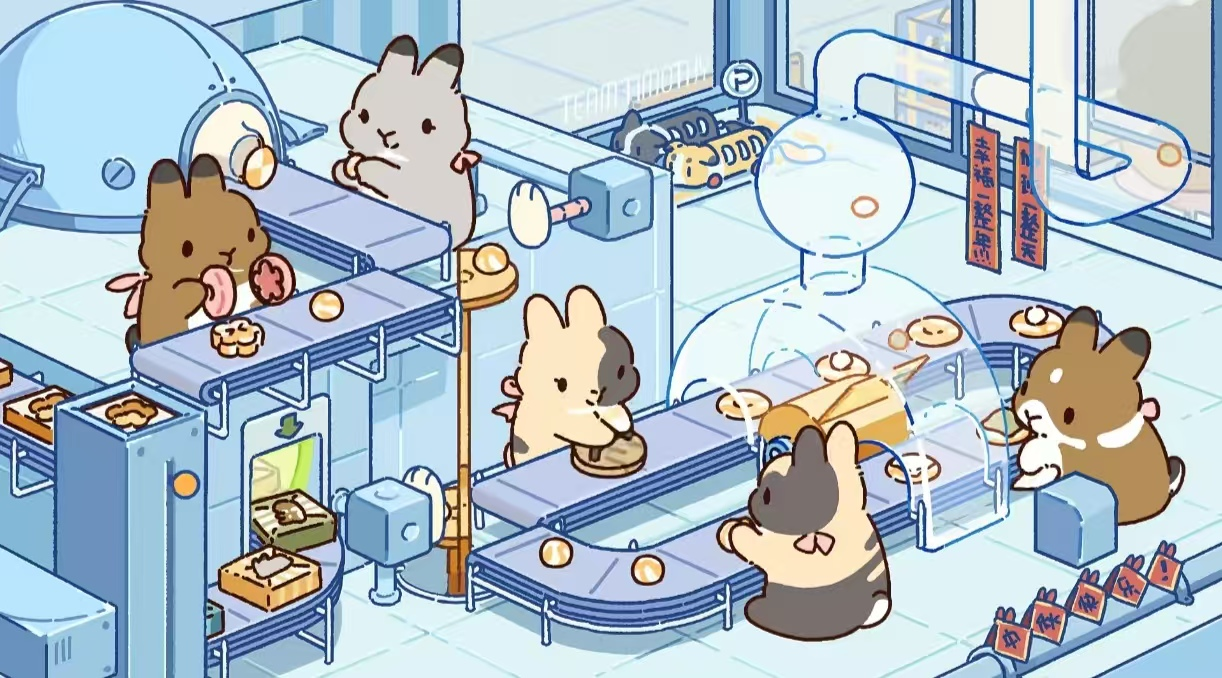} &
    \sampleobjects{
        \maskimg{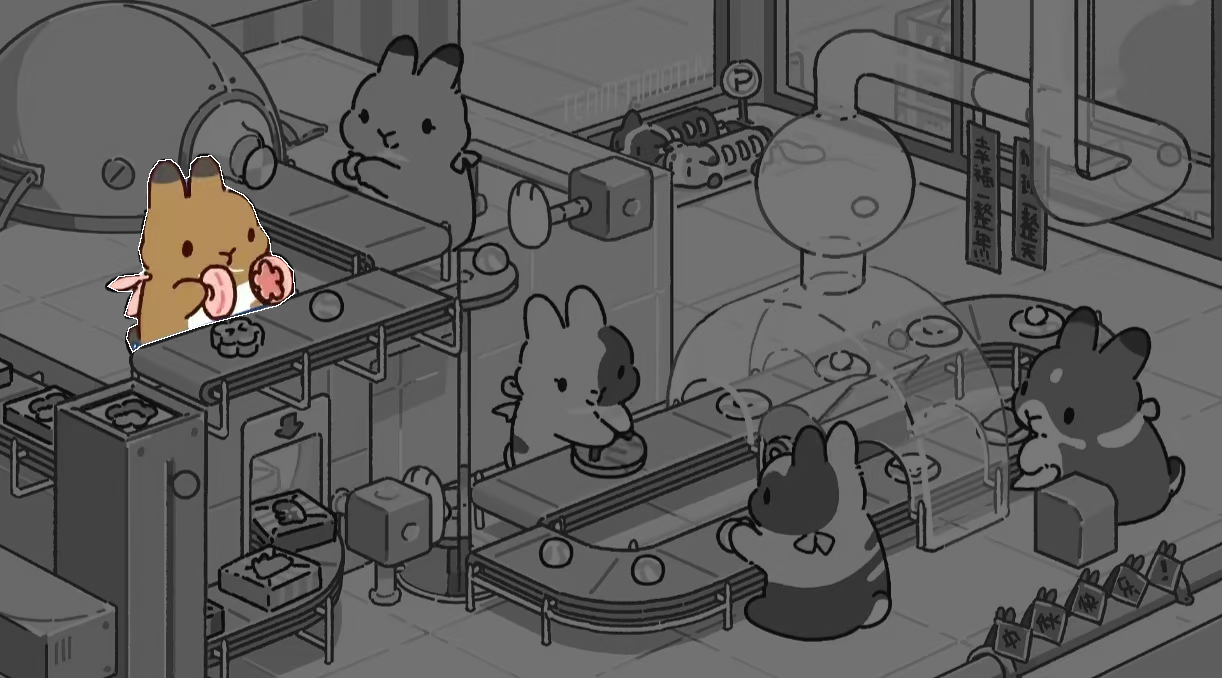} &
        Sanli &
        Please find the youngest one among the five rabbits in the image. \\
        \cmidrule(l){1-3}
        \maskimg{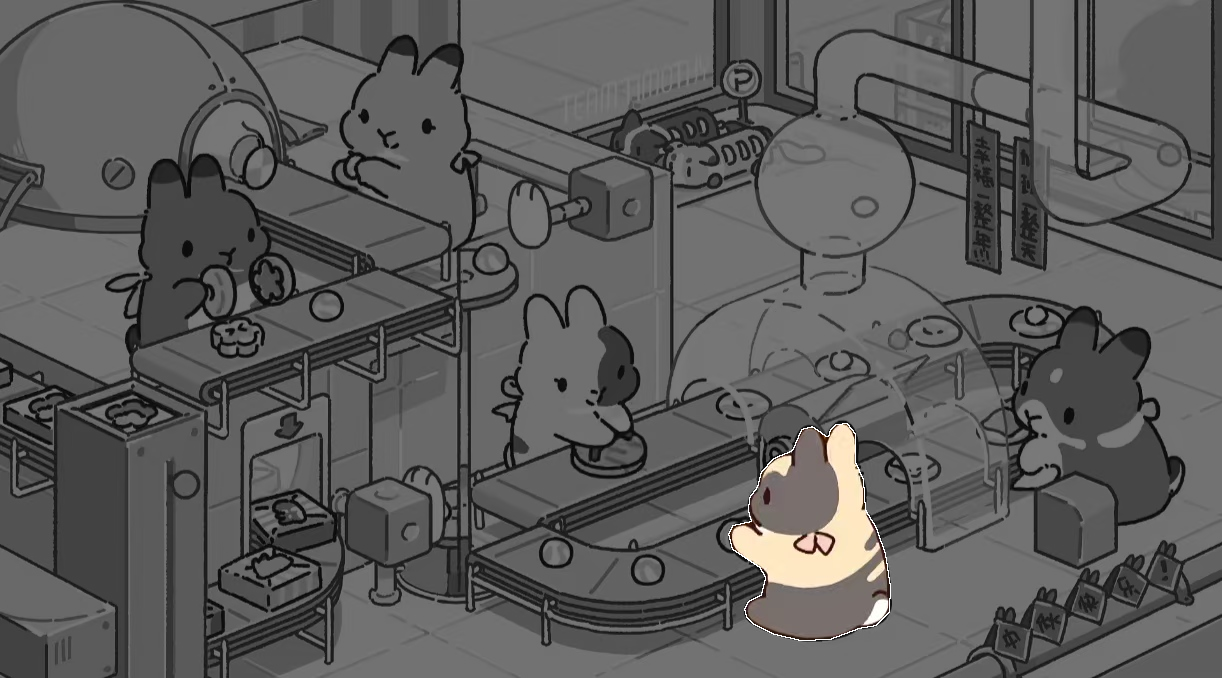} &
        Erli &
        Please find the youngest female rabbit in the image. \\
        \cmidrule(l){1-3}
        \maskimg{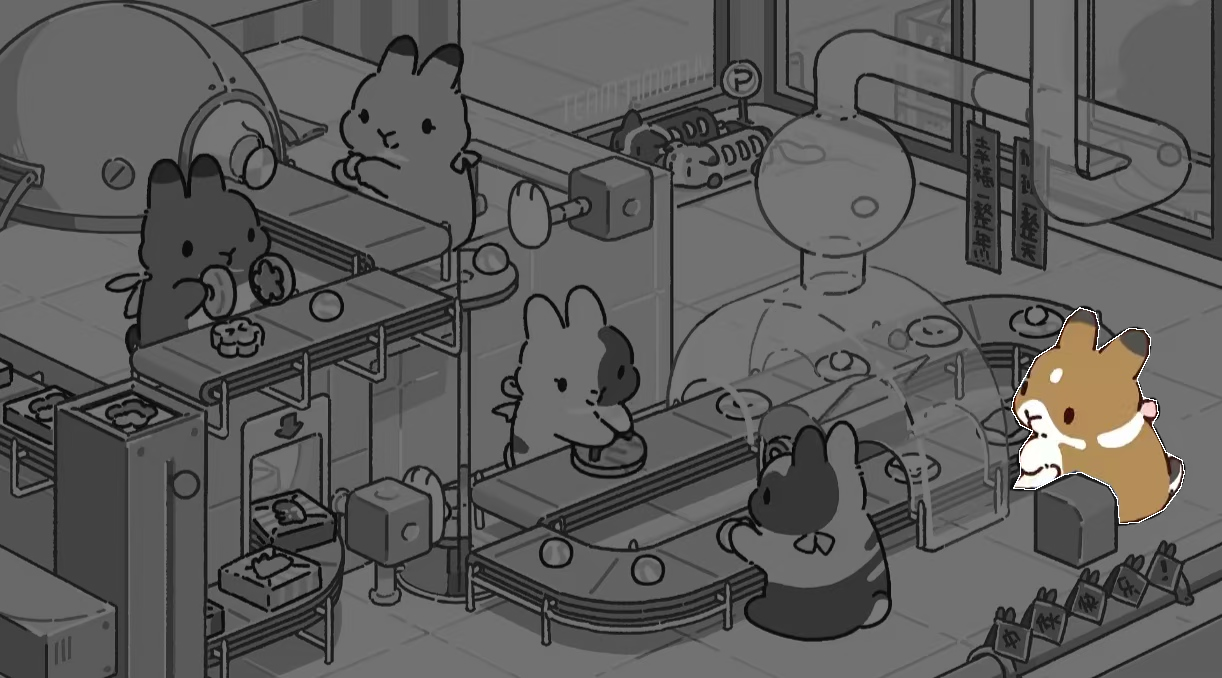} &
        Shidan &
        Please find the rabbit that was paired with Clarence during the collaboration between Lovebrush Chronicles and this IP in the image.
    } \\
    \midrule

    \sourcecell{0.150\textheight}{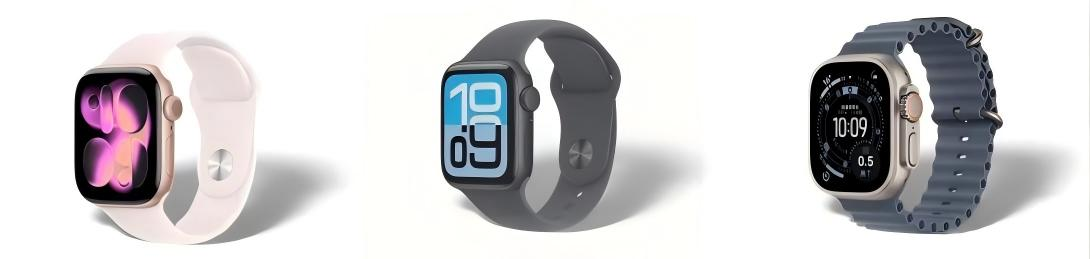} &
    \sampleobjects{
        \maskimg{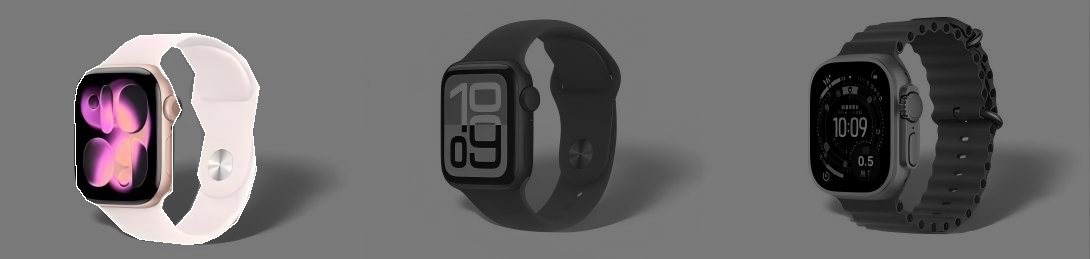} &
        Apple Watch Series 11 &
        Please find the product that comes in both 42mm and 46mm sizes in the image. \\
        \cmidrule(l){1-3}
        \maskimg{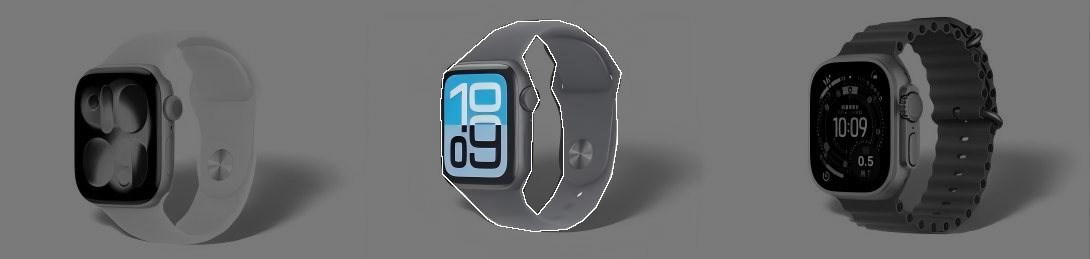} &
        Apple Watch SE 3 &
        Please find the product that uses only one specific metal material for the frame, and there are no other metal versions for this model, and this frame material is the same as the frame material of the Galaxy S26 Ultra in the image.
    } \\
    \midrule

    \sourcecell{0.235\textheight}{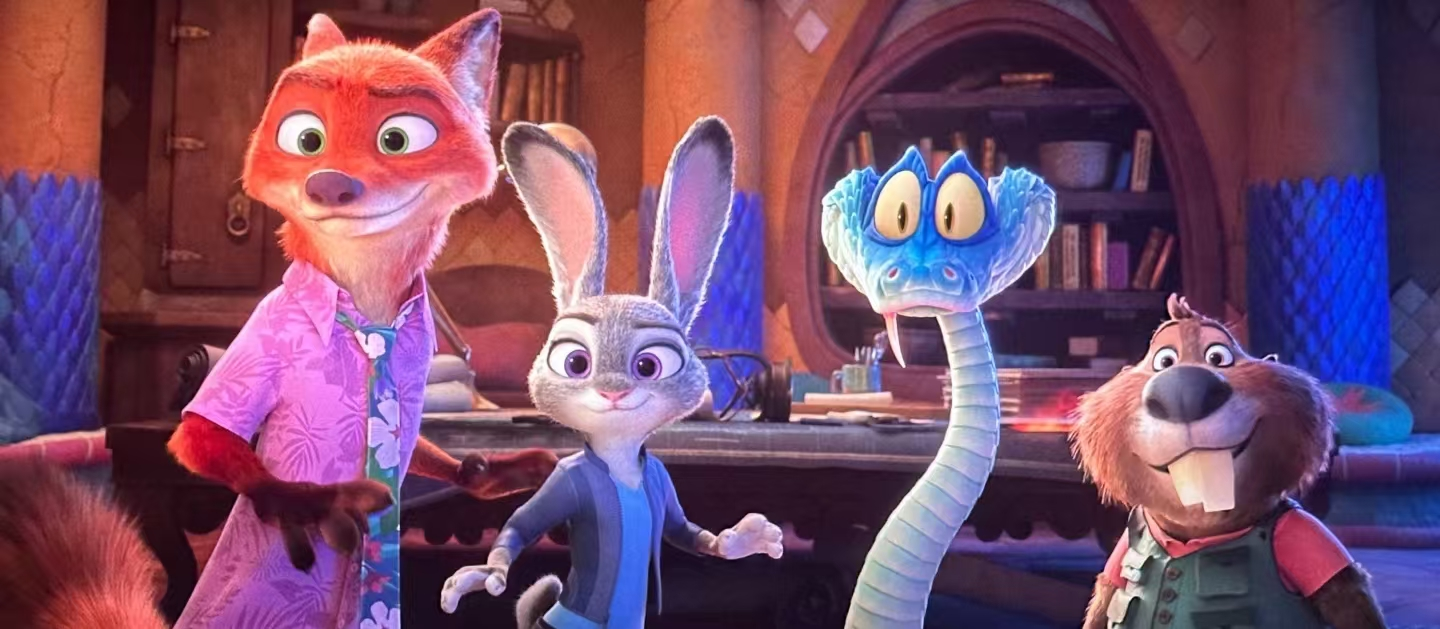} &
    \sampleobjects{
        \maskimg{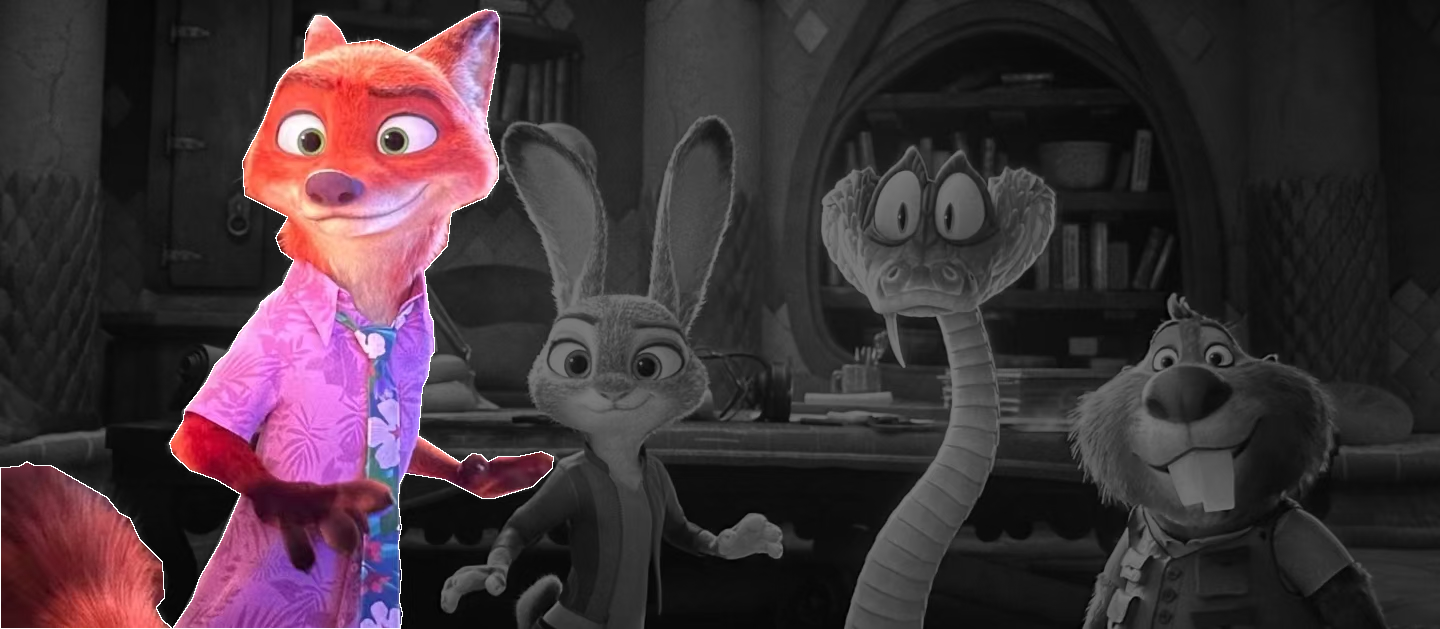} &
        Nick &
        Please find the character who knocked a snake unconscious with a frying pan in the film in the image. \\
        \cmidrule(l){1-3}
        \maskimg{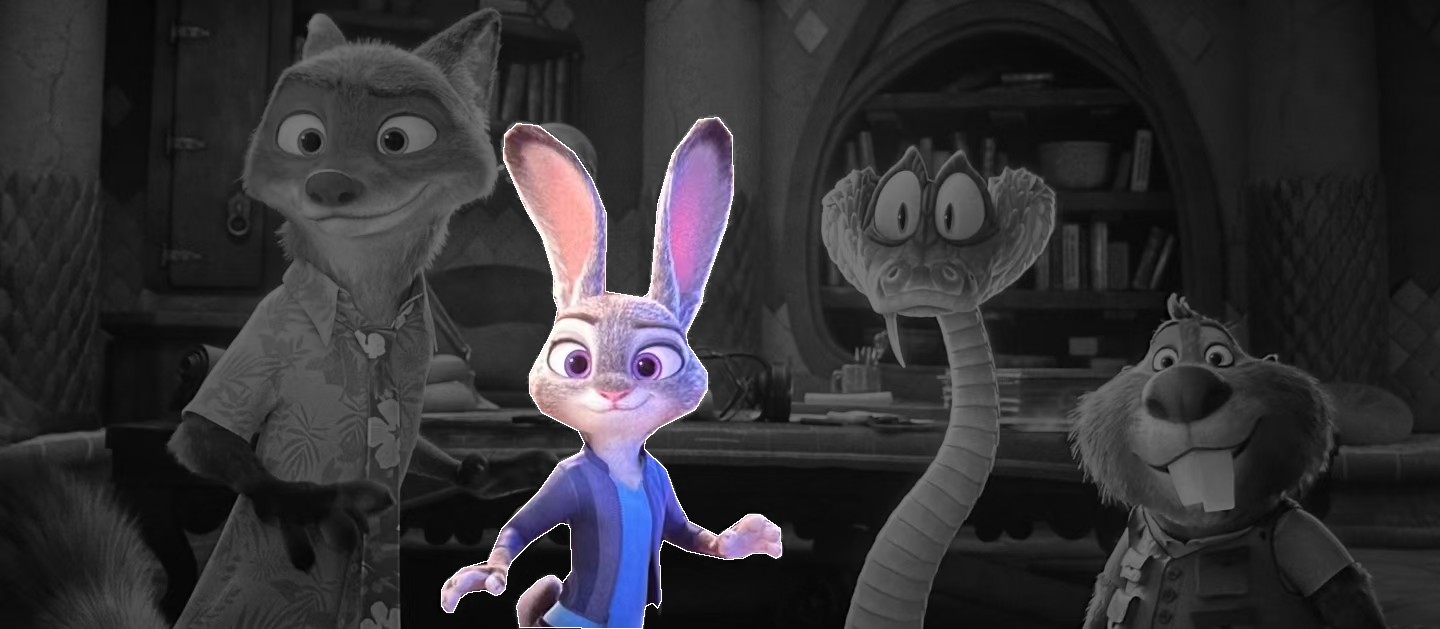} &
        Judy &
        Please find the character who was saved only by an antivenom pen being stabbed directly into the heart in the film in the image. \\
        \cmidrule(l){1-3}
        \maskimg{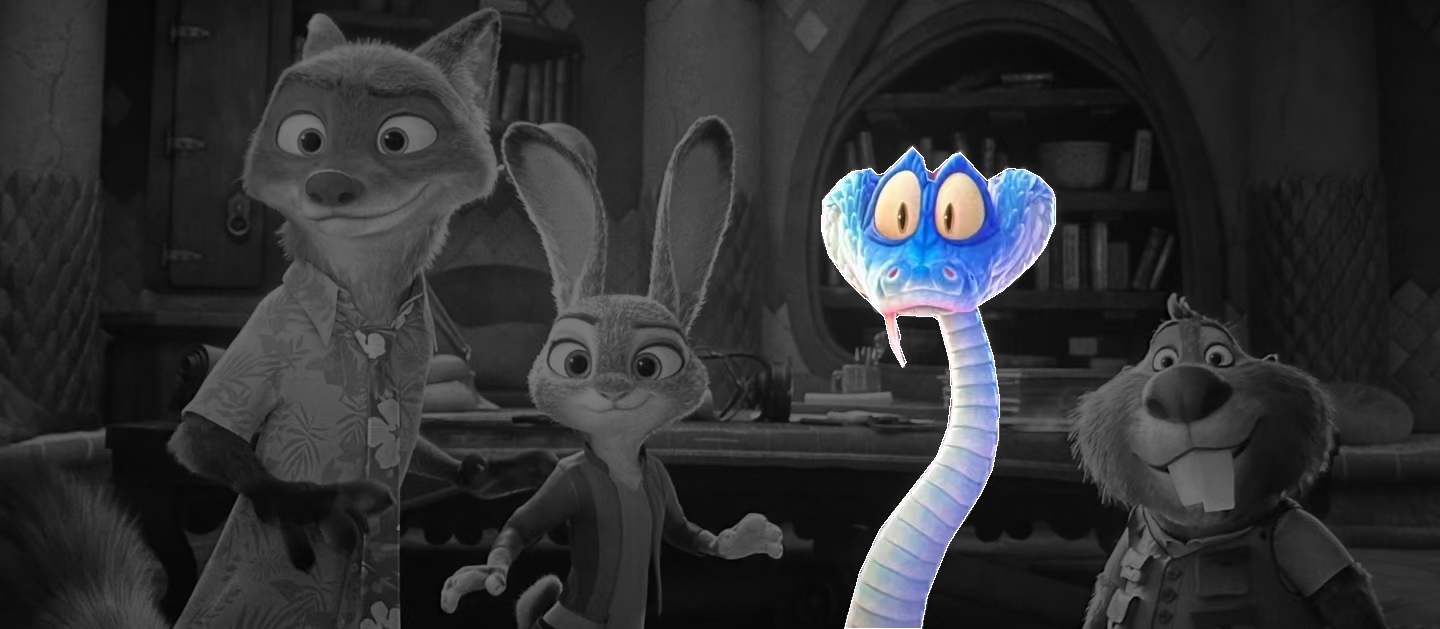} &
        Gary &
        Please find the character who revealed the map hidden beneath the metal cover of the Lynxley Journal in the film in the image. \\
        \cmidrule(l){1-3}
        \maskimg{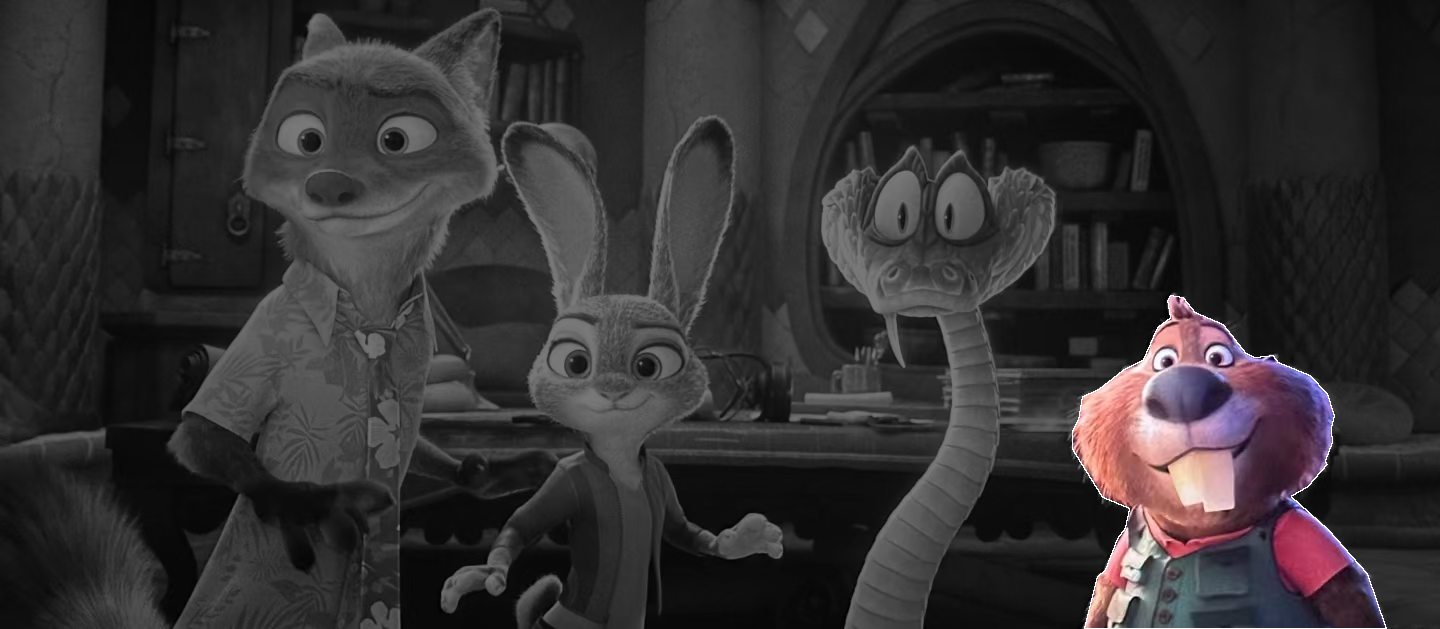} &
        Libao &
        Please find the character referred to as the ``Zootopia know-it-all'' in the film in the image.
    } \\
    \midrule

    \sourcecell{0.285\textheight}{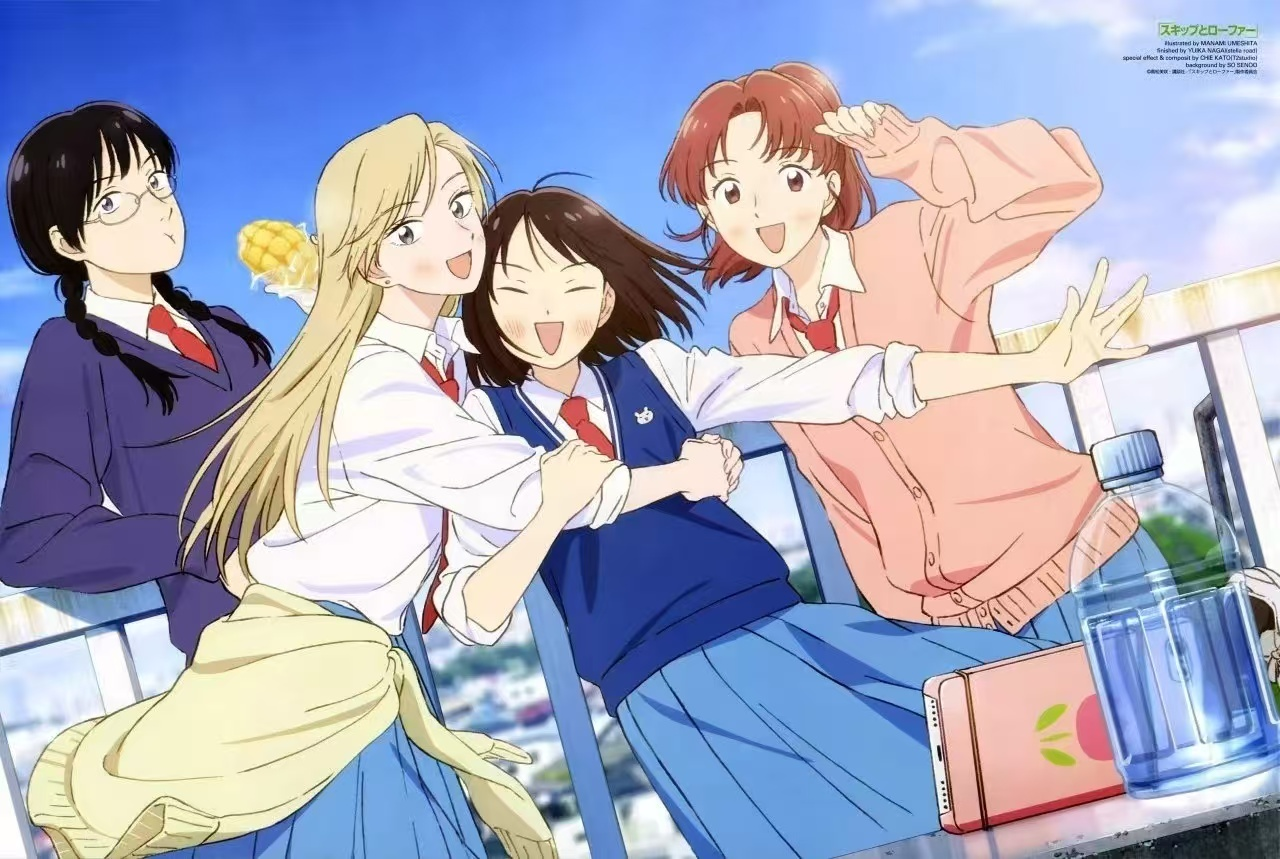} &
    \sampleobjects{
        \maskimg{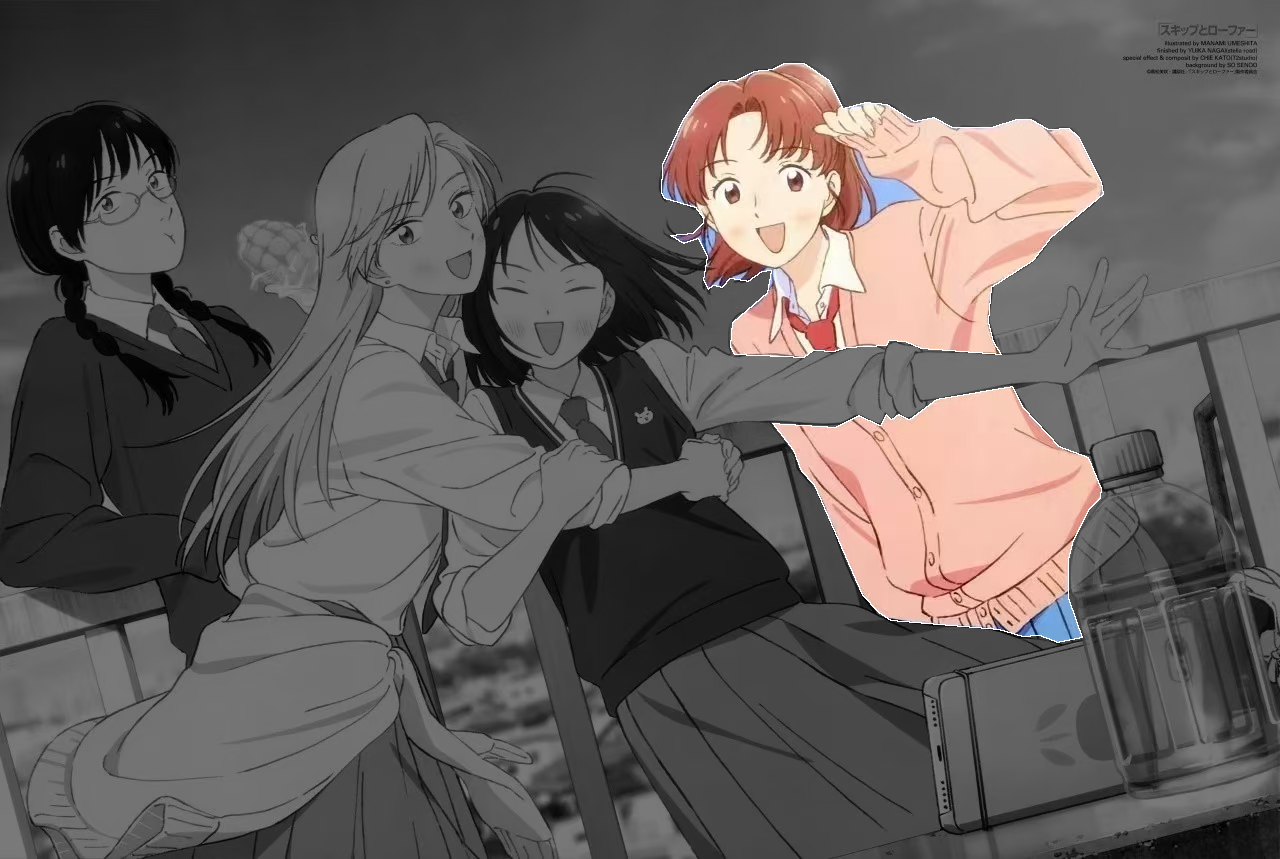} &
        Mika Egashira &
        Please find the character whose voice actor also voiced the character in Pokemon Horizons who holds the mysterious Poke Ball that can summon the black Rayquaza in the image. \\
        \cmidrule(l){1-3}
        \maskimg{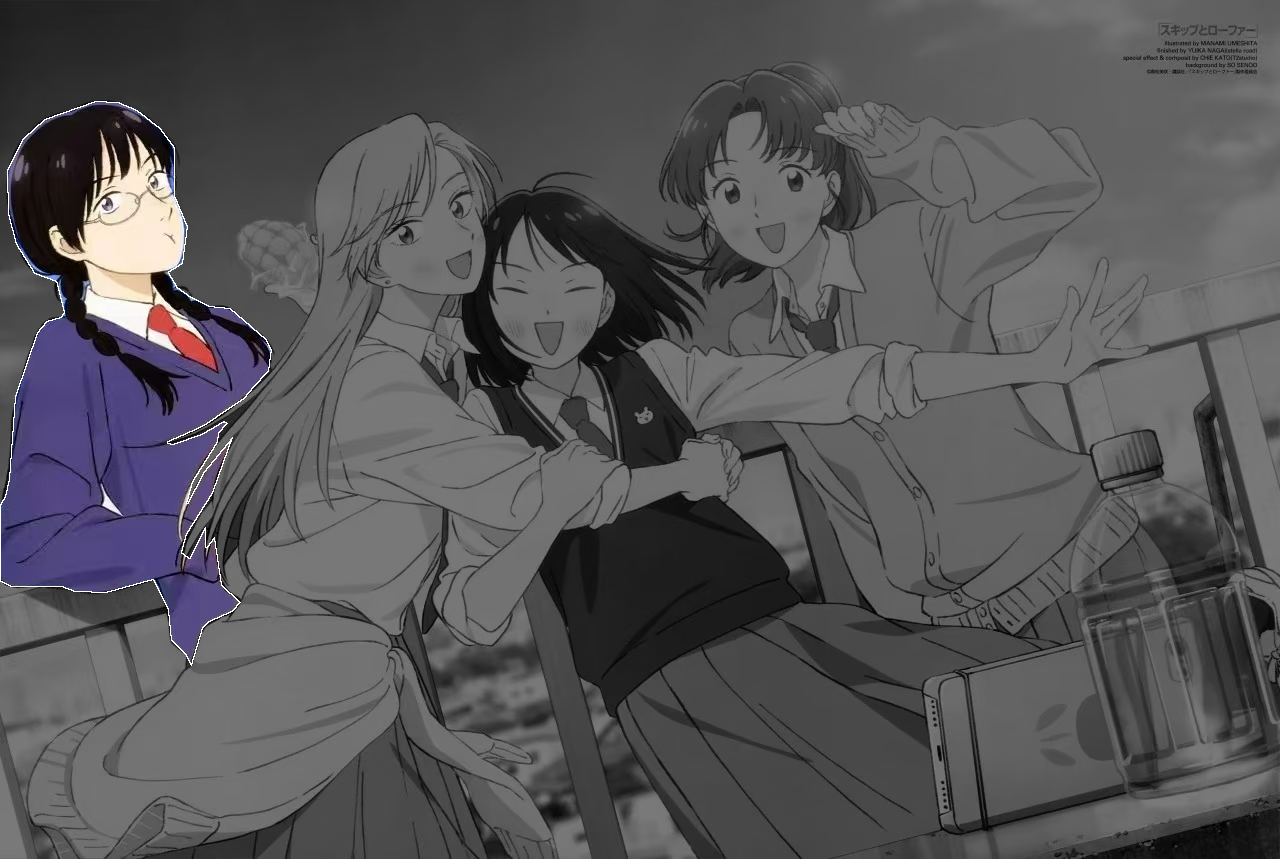} &
        Makoto Kurume &
        Please find the character whose voice actor also participated in the voice cast of the anime series directed by Daisuke Hiramaki that began airing in January 2026 in the image. \\
        \cmidrule(l){1-3}
        \maskimg{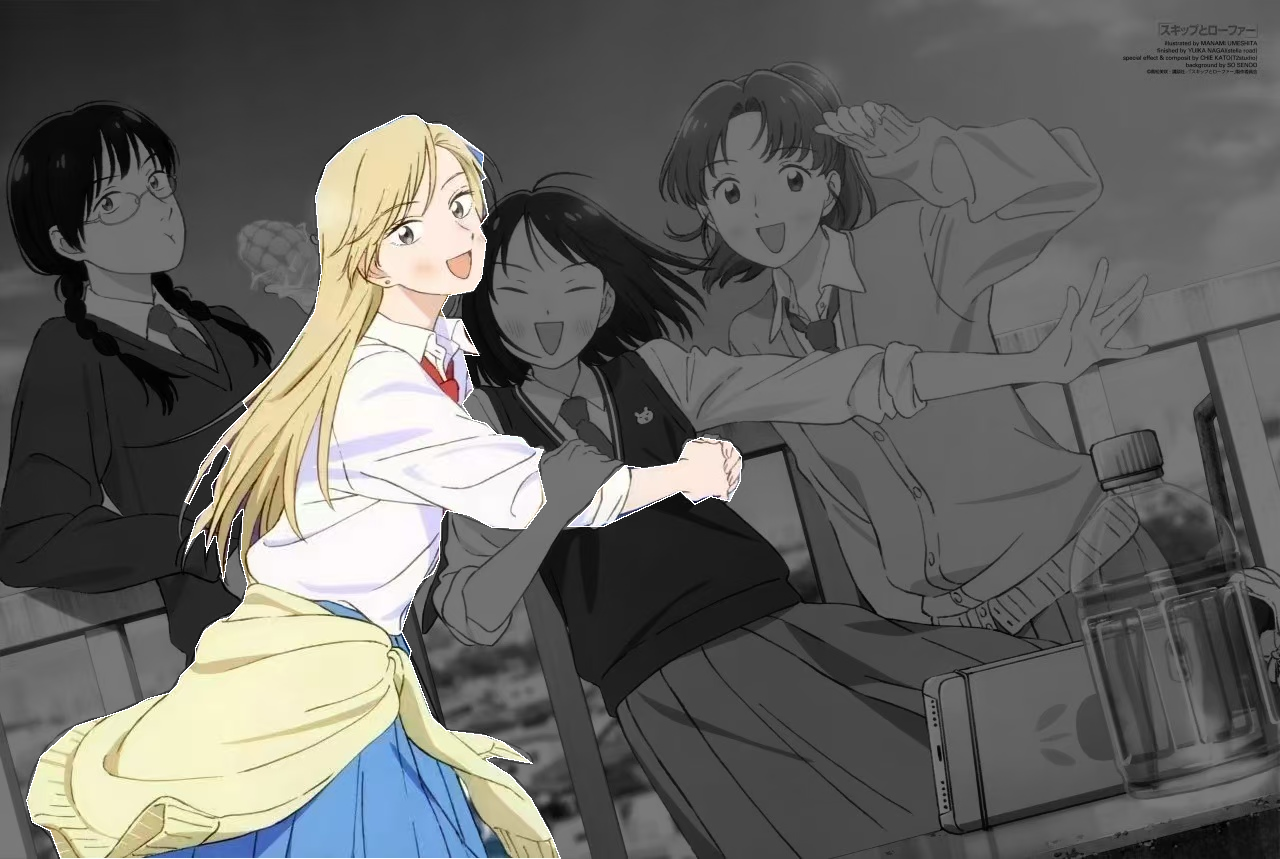} &
        Yuzuki Murashige &
        Please find the character who owns a dog named Chiffon in the image. \\
        \cmidrule(l){1-3}
        \maskimg{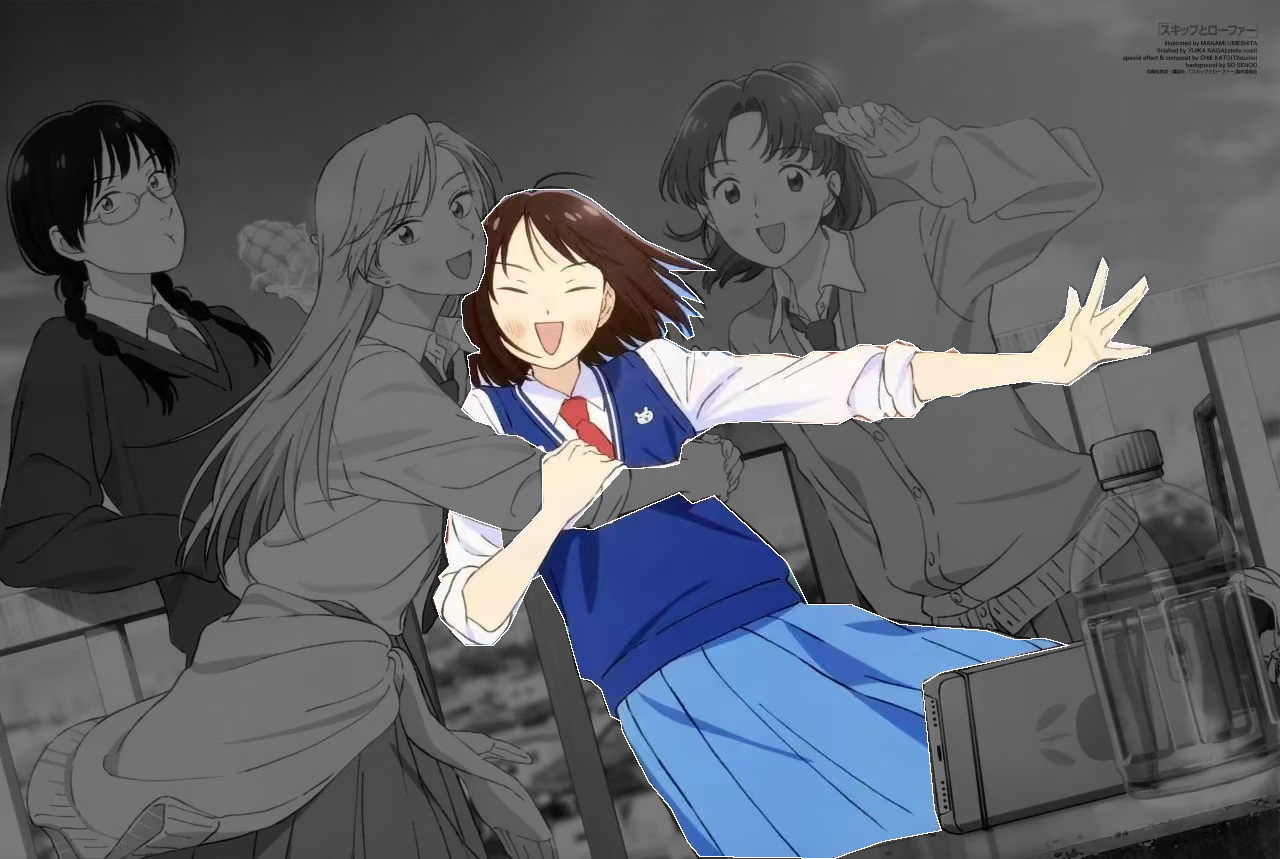} &
        Mitsumi Iwakura &
        Please find the character played by Miisha Shimizu in the musical Skip and Loafer, which premiered in March 2026, in the image.
    } \\
\end{longtable}
\endgroup

\newpage
\section{Prompt Templates}
\label{app:prompts}

This appendix lists the main prompt templates used by Pixel-Searcher. The placeholders in braces are filled with the sample-specific question, evidence, entity hypothesis, visual candidates, or answer options at inference time.

\begingroup
\setlength{\parskip}{0.15em}
\setlength{\parindent}{0pt}

\subsection{Hidden-Entity Search}

\begin{promptbox}{Question decomposition.}
You are decomposing a multi-hop visual grounding question into simpler atomic sub-questions that can each be answered by a single web search.

Question: {question}

Return strict JSON only:
{"sub_questions": ["sub-question 1", "sub-question 2", ...]}

Rules:
1. 1-3 sub-questions, ordered by reasoning dependency.
2. Each sub-question should target one hop of reasoning.
3. If the question is already simple, return it as the only sub-question.
4. Preserve the final target of the original question. If the question asks about the item/person in the image, the last sub-question must still ask about that final target, not about an intermediate clue.
5. Do not let a year, event, or historical clue replace the final grounded entity. Intermediate clues are for resolving the target, not for becoming the target.
6. Return only JSON.
\end{promptbox}

\begin{promptbox}{Multi-round search agent.}
You are a multi-round reasoning agent for visual grounding. Your goal is to identify the exact entity described by the question so it can be located in an image.

Original question: {question}
Sub-questions: {sub_questions}

Accumulated evidence so far:
{evidence}

Interaction round {round_num} of {max_rounds}. Only SEARCH / ANSWER consume an interaction round. THINK does not.

Return strict JSON with ONE of these forms:
1. {"action": "SEARCH", "query": "your DuckDuckGo search query"}
2. {"action": "THINK", "reasoning": "your reasoning based on evidence so far"}
3. {"action": "ANSWER", "entity_name": "resolved entity", "visual_category": "phone/person/car/...", "entity_type": "device/person/character/vehicle/object", "key_cues": ["cue1", "cue2"], "confidence": 0.0-1.0}

Guidelines:
- Use SEARCH to gather information you don't have yet.
- Use THINK only to briefly consolidate evidence before the next action.
- If evidence is still missing or ambiguous, prefer SEARCH over repeated THINK.
- You may use at most one THINK before you must SEARCH with a different query or ANSWER.
- Do not repeat the same reasoning across rounds.
- Use ANSWER only when you are confident about the entity.
- If ambiguity remains, issue a more targeted SEARCH instead of restating the same conclusion.
- If this is interaction round {max_rounds}, you MUST use ANSWER.
- Return only JSON.
\end{promptbox}

\begin{promptbox}{Forced answer.}
You have gathered the following evidence about this question. You must now give your best answer.

Question: {question}
Evidence: {evidence}

Return strict JSON:
{"entity_name": "best guess entity", "visual_category": "phone/person/car/...", "entity_type": "device/person/character/vehicle/object", "key_cues": ["cue1", "cue2"], "confidence": 0.0-1.0, "remaining_ambiguities": ["ambiguity1", "ambiguity2"]}

Rules:
- Return only JSON.
- Before answering, identify unresolved ambiguities from the evidence.
- If the evidence already resolves the entity, remaining_ambiguities can be [].
- Give your best guess even if uncertain.
\end{promptbox}

\begin{promptbox}{Final target resolution.}
Resolve the FINAL visible target of this grounding question.

Question: {question}
Evidence:
{evidence}

Return strict JSON:
{"entity_name": "final visible target", "visual_category": "phone/person/car/object/...", "entity_type": "device/person/character/vehicle/object", "key_cues": ["cue1", "cue2"], "confidence": 0.0-1.0}

Rules:
1. Answer the actual item/person that should be located in the image.
2. Do not answer with an intermediate clue entity, historical reference, designer, event, or source article unless that is also the visible target.
3. Prefer the concrete visible model/person/character over a generic series or franchise name.
4. Only return an exact model/person if the evidence explicitly supports that exact target; otherwise return the best supported visible target.
5. Return only JSON.
\end{promptbox}

\begin{promptbox}{Entity verification.}
You are checking whether a resolved entity is actually consistent with a visual grounding question and the gathered evidence.

Question: {question}
Proposed entity: {entity_name}
Visual category: {visual_category}
Entity type: {entity_type}
Key cues: {key_cues}

Evidence:
{evidence}

Return strict JSON:
{"is_consistent": true, "consistency_score": 0.0-5.0, "issues": ["issue 1", "issue 2"], "followup_queries": ["query 1", "query 2"]}

Rules:
1. Mark is_consistent false if the proposed entity seems to be the wrong product/person/character/model, too generic, unsupported by evidence, or an intermediate clue rather than the final visible target in the image.
2. consistency_score 5 means the entity is well supported and specific.
3. If inconsistent, provide 1-2 targeted followup_queries to resolve the remaining ambiguity.
4. For model-level answers, exact evidence matters. Do not mark an entity consistent unless the evidence explicitly supports that exact model/person, not just a nearby series, sibling model, platform, or speculative variant.
5. Return only JSON.
\end{promptbox}

\begin{promptbox}{Entity repair.}
The current resolved entity for a visual grounding question appears unreliable.

Question: {question}
Current entity: {entity_name}
Known issues with the current entity:
{issues}

Evidence:
{evidence}

Return strict JSON:
{"entity_name": "better entity", "visual_category": "phone/person/car/object/...", "entity_type": "device/person/character/vehicle/object", "key_cues": ["cue1", "cue2"], "confidence": 0.0-1.0}

Rules:
1. Re-resolve the entity from the evidence. Do not stick to the current entity if it is unsupported.
2. Prefer the most concrete model/person/character/entity actually supported by the evidence.
3. If the question asks about the item/person in the image, answer that final visible target, not an intermediate clue used to identify it.
4. Prefer an exact model/person only when it is explicitly supported by the evidence; otherwise step back to the best supported visible target.
5. If evidence is insufficient, still return the best alternative guess.
6. Return only JSON.
\end{promptbox}

\subsection{Visual Grounding}

\begin{promptbox}{Visual appearance extraction.}
Given search results about the appearance of "{entity_name}", extract a concise visual description focusing on shape, color, size, logos, and distinguishing physical features.

Search results:
{search_evidence}

Return strict JSON:
{"visual_description": "1-3 sentence description of how it looks", "shape": "compact/tall/flat/...", "color": "primary color(s)", "distinctive_features": ["feature1", "feature2"]}

Rules: Return only JSON. Focus on external appearance.
\end{promptbox}

\begin{promptbox}{Direct grounding.}
You are locating a specific entity in the FIRST image.

The FIRST image is the target scene. Any additional images are web reference images of the target entity.

Question: {reference_text}
Entity name: {entity_name}
Visual category: {visual_category}
Key cues: {key_cues}

Visual appearance summary:
{visual_description}

Return strict JSON in one of these forms:
{"bbox": [x1, y1, x2, y2], "confidence": 0.0-1.0, "reason": "short reason"}
{"bbox": null, "confidence": 0.0-1.0, "reason": "short reason"}

Rules:
1. bbox must use absolute pixel coordinates in the FIRST image only.
2. Use the attached reference images to find the same object/person/icon/model.
3. If several similar instances exist, choose the one best matching the cues.
4. Return a tight box around one concrete instance only. Avoid broad boxes that cover multiple objects, large empty regions, or the center area between objects.
5. If no plausible match exists, return bbox null.
6. Return only JSON.
\end{promptbox}

\begin{promptbox}{Candidate joint ranking.}
You are selecting the best matching candidate in the FIRST image.

The FIRST image is the full scene with all candidate boxes labeled.
The next candidate images are crops in this order:
{candidate_order}

Any remaining images are web reference images for the target entity.

Question: {reference_text}
Entity name: {entity_name}
Visual category: {visual_category}
Key cues: {key_cues}

Visual appearance summary:
{visual_description}

Candidates:
{candidate_lines}

Return strict JSON:
{"best_candidate_id": "candidate_x", "runner_up_candidate_id": "candidate_y or ''", "confidence": 0.0-1.0, "reason": "short reason"}

Rules:
1. Compare the labeled boxes in the overview image and the candidate crops jointly.
2. Prefer exact instance-level matches, not just same coarse category.
3. Use reference images when available.
4. Prefer the tighter candidate when two boxes cover the same object but one is broader or contains more background.
5. The reason must only cite visible evidence or explicit web evidence. Do not invent jersey numbers, logos, text, colors, or details that are not clearly visible.
6. If two candidates are similar, explicitly choose the better one.
7. Return only JSON.
\end{promptbox}

\begin{promptbox}{Candidate scoring.}
You are scoring whether a highlighted visual candidate matches a text hypothesis. You will see two target images first: the full image with the candidate highlighted in yellow, and a zoomed crop of the candidate region. You may also see additional web reference images of the entity after that.

Reference text: {reference_text}
Entity name: {entity_name}
Visual category: {visual_category}
Key cues: {key_cues}
Candidate id: {candidate_id}

Visual appearance from web search:
{visual_description}

Return strict JSON:
{"support_score": 0-5, "contradiction_score": 0-5, "confidence": 0.0-1.0, "reason": "short reason"}

Rules:
1. support_score: how much the visual evidence supports this being the entity.
2. contradiction_score: how much the visual evidence contradicts it.
3. Compare the candidate crop with the visual description and any attached reference images carefully.
4. Pay attention to shape, color, logos, text, layout, and distinctive features.
5. Broad boxes, multi-object boxes, and mostly-background boxes should receive lower support even if the right object is somewhere inside.
6. Only mention details that are actually visible or explicitly stated in the reference text. Do not invent hidden labels, numbers, or logos.
7. Return only JSON.
\end{promptbox}

\begin{promptbox}{Reference-image matching.}
You see two images:
1. A REFERENCE image of "{entity_name}" found on the web
2. A CANDIDATE crop from the target image

Does the candidate show the same type/model of object as the reference?

Return strict JSON:
{"match_score": 0-5, "reason": "short reason"}

Rules:
1. match_score 5 = definitely the same object type/model.
2. match_score 0 = completely different objects.
3. Focus on shape, color, brand logos, and distinctive features.
4. Ignore background differences; only compare the objects themselves.
5. Return only JSON.
\end{promptbox}

\begin{promptbox}{Visual entity repair.}
The current resolved entity does not visually match the objects found in the image. Re-resolve the target entity using both the question and the visible candidate summary.

Question: {question}
Current entity: {entity_name}
Current visual category: {visual_category}

Visible candidates in the image:
{candidate_summary}

Scoring evidence:
{score_summary}

Return strict JSON:
{"entity_name": "better entity", "visual_category": "phone/person/car/object/...", "entity_type": "device/person/character/vehicle/object", "key_cues": ["cue1", "cue2"], "confidence": 0.0-1.0}

Rules:
1. Use the visible candidates as a hard constraint: the answer should be an entity that could plausibly appear in this image.
2. Prefer specific model-level entities when the question implies them.
3. If the current entity is impossible given the visible candidates, replace it.
4. Return only JSON.
\end{promptbox}

\subsection{Grounding-Based VQA}

\begin{promptbox}{Option mini-resolution.}
Given this description of an entity, identify what it refers to.

Description: {text}

Return strict JSON:
{"entity_name": "identified entity", "visual_category": "phone/person/car/...", "entity_type": "device/person/character/vehicle/object", "key_cues": ["cue1", "cue2"], "confidence": 0.0-1.0}

Rules: Return only JSON.
\end{promptbox}

\begin{promptbox}{Grounded option selection.}
You see an image with one object highlighted in yellow. A cropped view of that object is also provided. Decide which option best describes this object.

Options:
{options_text}

Entity info from search:
{entity_info}

Return strict JSON:
{"selected_index": 0, "confidence": 0.0-1.0, "reason": "short reason"}

Rules:
1. selected_index is 0-based index of the best matching option.
2. Use visible evidence from the highlighted object.
3. Return only JSON.
\end{promptbox}

\subsection{Fallback Candidate Generation}

\begin{promptbox}{Object detection.}
You are a visual object detector. Identify ALL visually salient foreground objects in this image. For each object output a tight bounding box.

Return strict JSON only:
{"detections": [{"label": "short label", "bbox": [x1, y1, x2, y2]}]}

Rules:
1. bbox uses absolute pixel coordinates, format xyxy.
2. One entry per visible object instance.
3. Labels should be short and concrete, e.g. phone, person, car, router, laptop.
4. At most {max_boxes} boxes.
5. Boxes must be tight; do NOT include large background margins.
6. Do not output reasoning or markdown, only JSON.
\end{promptbox}

\begin{promptbox}{Saliency ranking.}
You are ranking detected objects by visual saliency (how visually prominent and attention-grabbing each object is).

The image contains these detected objects:
{candidate_list}

For each candidate, assign a saliency_score between 0.0 (background clutter) and 1.0 (most prominent foreground object).

Return strict JSON only:
{"scores": [{"id": "candidate_1", "saliency_score": 0.0}, ...]}

Rules:
1. Return only JSON.
2. Every candidate id must appear exactly once.
3. Do not output reasoning.
\end{promptbox}
\endgroup

\end{document}